\documentclass{article}

\usepackage{arxiv}          
\usepackage[T1]{fontenc}    
\usepackage{hyperref}       
\usepackage{url}            
\usepackage{booktabs}       
\usepackage{amsfonts}       
\usepackage{amsthm}         
\usepackage{nicefrac}       
\usepackage{microtype}      
\usepackage{graphicx}	      
\usepackage{enumitem}	      
\usepackage{tabularx}       
\usepackage{multirow}       
\usepackage{caption}        
\usepackage{siunitx}        
\usepackage{placeins}       
\usepackage[lofdepth,lotdepth]{subfig}
\usepackage[style=numeric,sorting=none]{biblatex}

\date{} 
\renewcommand{\undertitle}{} 
\renewcommand{\headeright}{} 

\newcolumntype{L}{>{\raggedright\arraybackslash}X}
\newcolumntype{C}{>{\centering\arraybackslash}X}
\newcolumntype{R}{>{\raggedleft\arraybackslash}X}
\newcolumntype{O}[1]{>{\raggedright\arraybackslash}p{#1}}
\newcolumntype{P}[1]{>{\centering\arraybackslash}p{#1}}
\newcolumntype{Q}[1]{>{\raggedleft\arraybackslash}p{#1}}

\theoremstyle{definition}
\newtheorem{definition}{Definition}

\hypersetup{
    colorlinks=true,
    urlcolor=blue,
    pdftitle={Towards an Analytical Definition of Sufficient Data},
    pdfpagemode=FullScreen,
}

\title{Towards an Analytical Definition of Sufficient Data}

\author{
  Adam~Byerly\\
  Department of Electronic and Electrical Engineering\\
  Brunel University London\\
  Uxbridge, UB8 3PH UK \\
  Department of Computer Science and Information Systems\\
  Bradley University\\
  Peoria, Il, 61615 USA\\
  \texttt{abyerly@fsmail.bradley.edu} \\
  \And{}
  Tatiana~Kalganova \\
  Department of Electronic and Electrical Engineering\\
  Brunel University London\\
  Uxbridge, UB8 3PH UK \\
  \texttt{tatiana.kalganova@brunel.ac.uk} \\
}

\begin{document}

\maketitle

\begin{abstract}
We show that, for each of five datasets of increasing complexity, certain training samples are more informative of class membership than others.  These samples can be identified \textit{a priori} to training by analyzing their position in reduced dimensional space relative to the classes' centroids.  Specifically, we demonstrate that samples nearer the classes' centroids are less informative than those that are furthest from it.  For all five datasets, we show that there is no statistically significant difference between training on the entire training set and when excluding up to 2\% of the data nearest to each class's centroid.
\end{abstract}


\section{Introduction}\label{sec:introduction}

The experimental results of~\cite{Byerly2021c} demonstrated that for the micro-PCB dataset they introduced, a portion of the data could be excluded from training and models could achieve comparable accuracy when using data augmentation to simulate the particular attributes of the excluded data, particularly when those models used Homogeneous Vector Capsules~\cite{Byerly2021a}.  The micro-PCB dataset consists of images coded for rotation and perspective and that allowed for data augmentation techniques to simulate known excluded rotations and perspectives.  Since most image classification datasets aren't coded in any way beyond class membership, the question naturally arises:  When no coding is available, is there a metric that can be extracted directly from the image data that can be acted upon in a useful way?

Since the primary goal in classification is to distinguish between classes, finding metrics that are indicative of those classes, or more specifically, indicative of those classes relative to the other classes, is a corollary goal.  The hypothesis that is tested here is that there exists a distance metric that can be leveraged during the training of a convolutional neural network in order to improve or at least maintain testing accuracy by using that metric to \textit{exclude} a portion of the training data.  

\begin{figure}[!htbp]
  \begin{center}
  \setlength\tabcolsep{3pt}
  \begin{tabular}{@{}cccccccccc@{}}
    \includegraphics[width=0.5in]{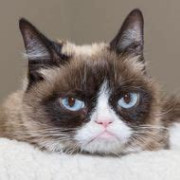} &
    \includegraphics[width=0.5in]{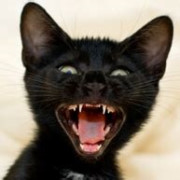} &
    \includegraphics[width=0.5in]{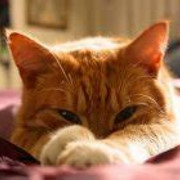} &
    \includegraphics[width=0.5in]{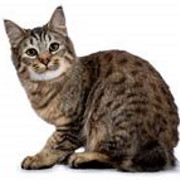} &
    \includegraphics[width=0.5in]{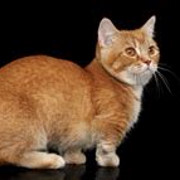} &
    \includegraphics[width=0.5in]{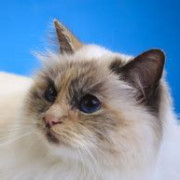} &
    \includegraphics[width=0.5in]{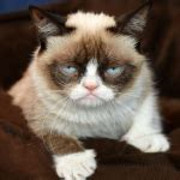} &
    \includegraphics[width=0.5in]{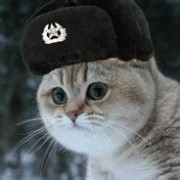} &
    \includegraphics[width=0.5in]{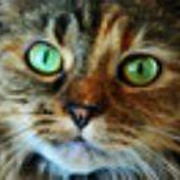} &
    \includegraphics[width=0.5in]{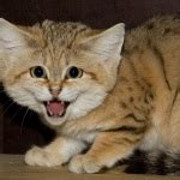} \\
    [.1 in]
  \end{tabular}
  \begin{tabular}{@{}cccccccccc@{}}
    \includegraphics[width=0.5in]{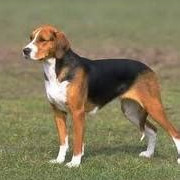} &
    \includegraphics[width=0.5in]{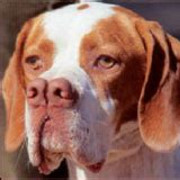} &
    \includegraphics[width=0.5in]{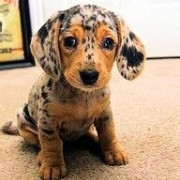} &
    \includegraphics[width=0.5in]{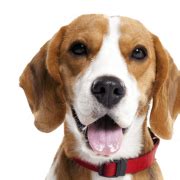} &
    \includegraphics[width=0.5in]{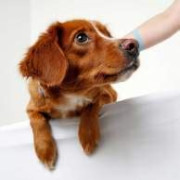} &
    \includegraphics[width=0.5in]{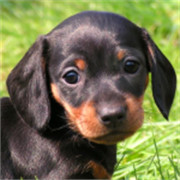} &
    \includegraphics[width=0.5in]{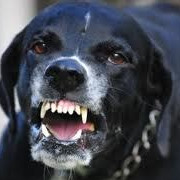} &
    \includegraphics[width=0.5in]{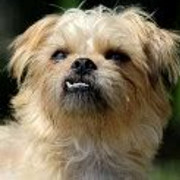} &
    \includegraphics[width=0.5in]{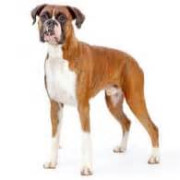} &
    \includegraphics[width=0.5in]{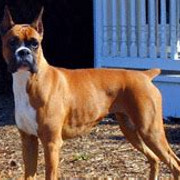} \\
    [.1 in]
  \end{tabular}
  \begin{tabular}{@{}cccccccccc@{}}
    \includegraphics[width=0.5in]{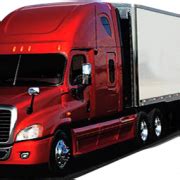} &
    \includegraphics[width=0.5in]{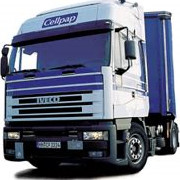} &
    \includegraphics[width=0.5in]{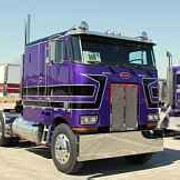} &
    \includegraphics[width=0.5in]{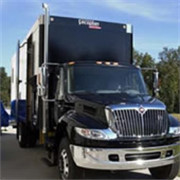} &
    \includegraphics[width=0.5in]{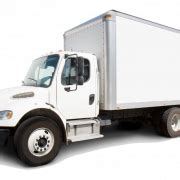} &
    \includegraphics[width=0.5in]{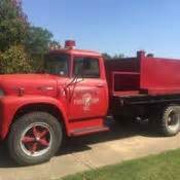} &
    \includegraphics[width=0.5in]{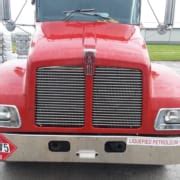} &
    \includegraphics[width=0.5in]{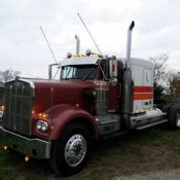} &
    \includegraphics[width=0.5in]{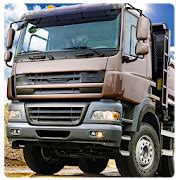} &
    \includegraphics[width=0.5in]{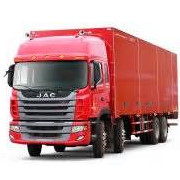} \\
  \end{tabular}
\end{center}
\caption{A Small 3-Class Dataset Consisting of Cats, Dogs, and Trucks}\label{fig:3class_demo_images}
\end{figure}

\begin{figure}[!htbp]
  \centering
  \includegraphics[width=.98\textwidth]{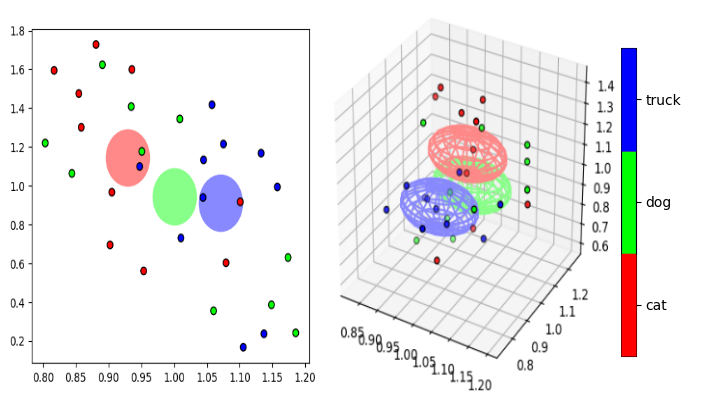}
  \caption{Visualization of 2-Dimensional and 3-Dimensional Reductions of the Cat, Dog, Truck Dataset.  The circles (in 2D) and spheres (in 3D) represent the classes' centroids.}\label{fig:3class_demo_2d_and_3d_reduction}
\end{figure}

An obvious candidate for a distance metric is Euclidian distance from the centroid for the class.  An obvious problem with Euclidian distance is the curse of dimensionality that results in near uniformity of distance from the centroid among samples when measuring distance in a large number of dimensions.  For example, take a small 3-class dataset, consisting of 10 images in each of the classes cat, dog, and truck derived from a Google Image search for each of the 3 terms ``cat'', ``dog'' and ``truck''  as shown in \autoref{fig:3class_demo_images}.  These images are 180 pixels square and made up of 3 color channels.  Thus, the dataset is 97,200-dimensional.  Fortunately, there exists a class of non-linear dimensionality reduction techniques that can learn to represent high-dimensional data in lower dimensions.  These include t-distributed stochastic neighbor embedding (t-SNE)~\cite{Maaten2008} and Uniform Manifold Approximation and Projection (UMAP)~\cite{McInnes2018}.  Due to its relative speed (as compared to t-SNE) and strong theoretical foundations, UMAP is quickly becoming one of the most popular non-linear general dimensionality reduction algorithms in use.  \autoref{fig:3class_demo_2d_and_3d_reduction} shows the result of using UMAP to reduce the dog, cat, truck dataset to 2 and 3 dimensions.  In this low-dimensional space the Euclidian distances for each sample from each class's centroid take on meaningful differences among themselves.

\begin{figure}[!htbp]
  \centering
  \includegraphics[width=.98\textwidth]{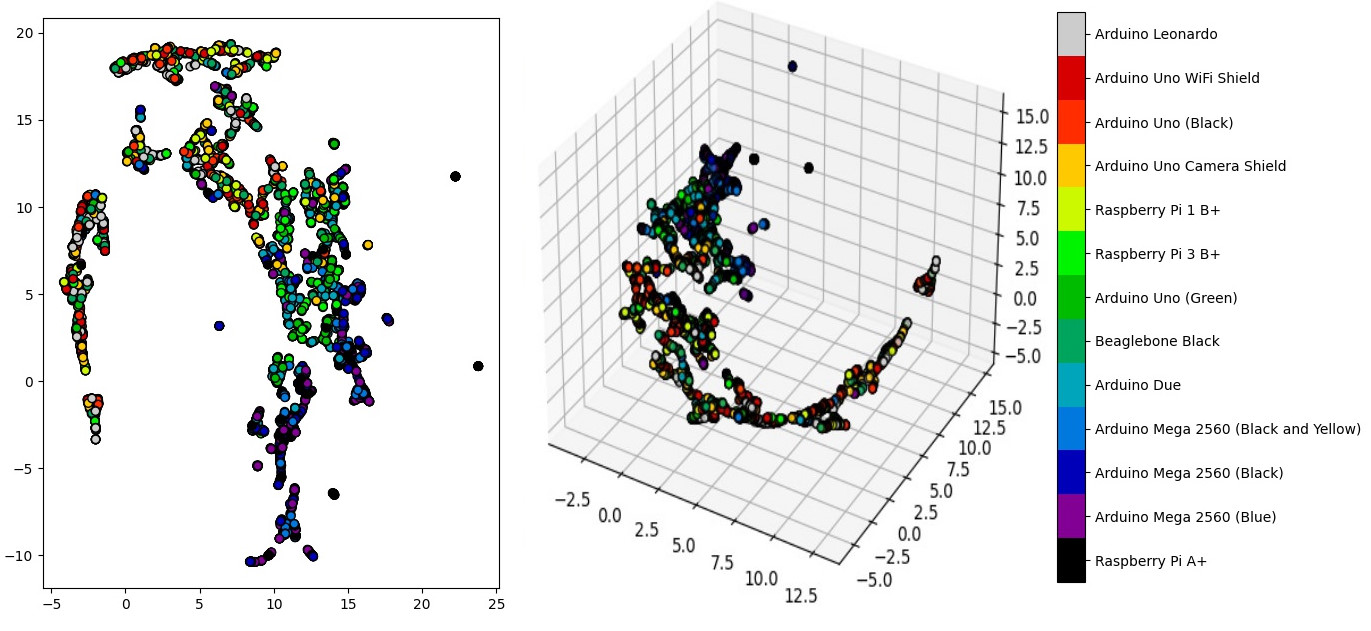}
  \caption{Visualization of 2-Dimensional and 3-Dimensional Reductions of the micro-PCB Dataset}\label{fig:micro_pcb_2d_and_3d}
\end{figure}

\begin{figure}[!htbp]
  \centering
  \includegraphics[width=.49\textwidth]{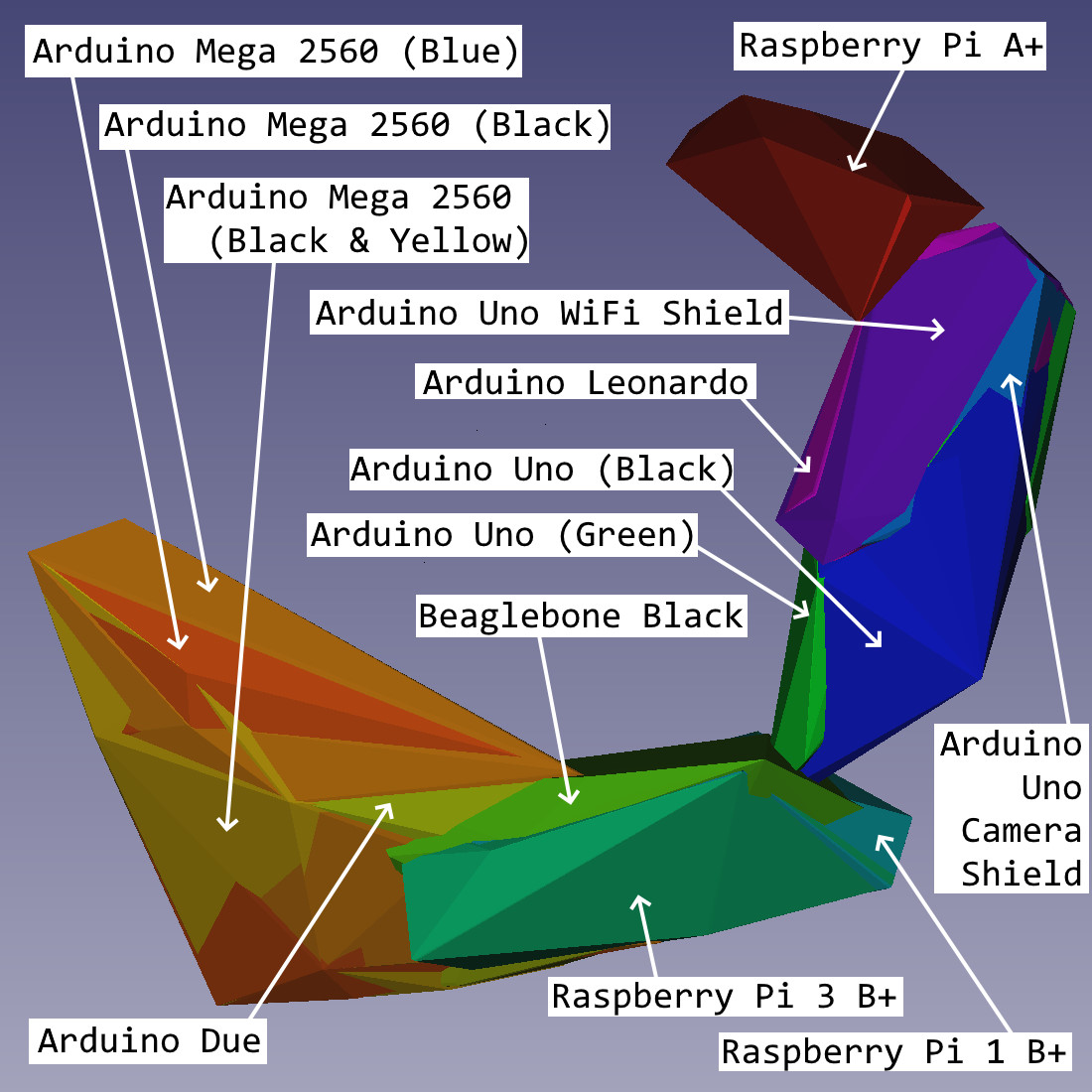}
  \caption{Visualization of 3-Dimensional Reduction of the micro-PCB Dataset (Excluding Outliers)}\label{fig:micro_pcb_3d}
\end{figure}

\autoref{fig:micro_pcb_2d_and_3d} shows 2 and 3-dimensional UMAP reductions of the micro-PCB training data.  When viewed as a collection of small epsilon balls around each point it is difficult to observe much useful structure.  Compare these with \autoref{fig:micro_pcb_3d}, which shows a 3-dimensional reduction of the micro-PCB training data produced with UMAP after removing the outliers from each class (those points more than one standard deviation away from their class's centroid) and enclosing the points in a convex hull.  Using this visualization method it is possible to observe interesting structure that is not apparent in the visualizations in \autoref{fig:micro_pcb_2d_and_3d}.  In the micro-PCB dataset there are three classes of micro-PCBs that are of the same model (Arduino Mega 2560) but produced by different manufacturers.  The boards' components and layouts are the same, differing almost entirely by the colors of the substrates and the colors of plastics used for the pins.  The reduction resulted in these classes being tightly clustered together and mostly separate from the remaining classes.  A similar tight clustering is exhibited with the two Arduino Unos created by different manufacturers, the two different generations of the Raspberry Pi B+, and the two different Arduino shields.

The ability for this 3-dimensional UMAP reduction to effectively separate the classes such that important semantic information of the micro-PCB dataset emerges in the locations of the samples in the reduced space serves as evidence that a distance metric can indeed be leveraged.

The experiments detailed in this work use the dimensional reduction of each sample as produced by UMAP to determine what samples to exclude during training.  These experiments are performed on standard benchmark datasets in order to make the results comparable to other research.  It should be noted that with all experiments detailed in this work, all exclusionary methods were applied only to the training set of each dataset and only during training.  All testing for all experiments used the entire test set of each dataset.

\subsection{Our Contribution}
Our contribution is as follows:
\begin{enumerate}
  \item We demonstrate that training samples in a reduced dimensional space furthest from their class's centroid are more informative than those samples located nearest to the class's centroid, thus providing an \textit{a prior} method for identifying the most informative samples.
  \item We demonstrate that reducing the training set size by 2\% for all datasets studied produced a statistically insignificant difference in accuracy as compared to when training on all samples.  The training set of CIFAR-10 was able to be reduced by 5\% and the training set for MNIST was able to be reduced by 10\%.
  \item We demonstrate that using a 3 dimensional reduction to calculate distances is sufficient relative to 2, 5, and 10 dimensional reductions for all datasets studied.
\end{enumerate}

\section{Related Work}\label{sec:related_work}

k-Nearest Neighbor (kNNs)~\cite{Cover1967} and other so called ``instance-based'' methods are also sometimes called ``memory-based'' because building the model from the training data involves storing the training instances.  During evaluation, the sample being evaluated is compared to each stored instance in order to find the best classification for it.  Thus, evaluation is an \(O(n)\) operation, where \(n\) is the number of stored instances.  Because the computation cost during evaluation is so high, these algorithms are also sometimes called ``lazy learners''.

Prior to the relatively recent pivot to large, over-parameterized neural networks for classification, these instance-based methods were some of the most widely-used for classification.  As such, a fair amount of research went into reducing the number of training samples that had to be ``memorized'' both to ease the memory requirements and to speed up evaluation.  Early methods for doing this include Condensed Nearest Neighbor (CNN)~\cite{Hart1968}, Selective Nearest Neighbor (SNN)~\cite{Ritter1975}, and Edited Nearest Neighbor (ENN)~\cite{Wilson1972}.  More recently a slew of other methods have been investigated~\cite{Wilson2000}\cite{Albalate2007}\cite{Vazquez2005}\cite{ChienHsing2006}\cite{Ougiaroglou2012}.

However, after the success of AlexNet~\cite{Krizhevsky2012}, research into classification methods pivoted to large, over-parameterized neural networks.  Shortly after that, some research into reducing training dataset size was conducted~\cite{Shayegan2014}, however, the focus was on acceptable loss of accuracy, rather than maintaining or improving accuracy.

More recently, the focus has shifted to adding training data beyond datasets' canonical training set in order to improve accuracy~\cite{Dosovitskiy2020}\cite{Kolesnikov2020}\cite{Touvron2020a}\cite{Pham2020}\cite{Xie2020}\cite{Foret2020}\cite{Zhai2021}\cite{Riquelme2021}\cite{Ryoo2021}\cite{Jia2021}\cite{Dong2021}\cite{Liu2021}\cite{Dai2021}\cite{Wu2021}\cite{Tan2021}\cite{Tolstikhin2021}\cite{Brock2021}.

\section{Network Architecture and Training}\label{network_architecture}

For all experiments, regardless of the dataset used:

\begin{enumerate}
    \item The network used for training consisted of a single set of stacked 3\(\times{}\)3 convolutions, wherein the first convolutional operation produced 32 feature maps.
    \item All subsequent convolutional operations in the network produced an additional 16 feature maps.
    \item After all convolutional layers in the network, a set of Unbroken Z-Derived Homogeneous Vector Capsules were used to produce the final classification.
    \item Optimization was performed with the Adam optimizer with an initial learning rate of 0.001 that was exponentially decayed every epoch by 0.98.
    \item Training proceeded for 300 epochs.
\end{enumerate}

However, because different datasets are formed from images with different sizes, differing number of color channels, and differing complexity of the features present, some slight differences were required depending on the dataset being trained on.

\subsection{Training on MNIST and Fashion-MNIST}

These datasets~\cite{Lecun2010}\cite{Xiao2017} are composed of single color-channel images comprised of the simplest features relative to the other datasets.  They all also use the smallest initial image size of 28\(\times{}\)28 pixels.  Thus, the network used on these datasets consisted of the fewest convolutional layers (9) and thus the fewest final set of feature maps (160).  No padding was used during the convolutional operations, so the final set of feature maps were 10\(\times{}\)10.  When training this network, a batch size of 120 was used.  Data augmentation used during the training for these datasets was uniform and the same strategy as described in~\cite{Byerly2021b}.

\subsection{Training on CIFAR-10 and CIFAR-100}

These datasets~\cite{Krizhevsky2009} are composed of 3 color-channel images comprised of more complex features than MNIST and Fashion-MNIST.\@ Additionally, the images are slightly larger at 32\(\times{}\)32 pixels.  By using a similar network as was used for MNIST and Fashion-MNIST, but with 2 more convolutional layers and thus 192 feature maps coming out of the final layer, the final set of feature maps were also 10\(\times{}\)10.  As in the case with MNIST and Fashion-MNIST, when training this network, a batch size of 120 was used.  Data augmentation used during the training for these datasets was uniform and the same strategy as used for the experiments in~\cite{Byerly2021a}.

\subsection{Training on Imagenette}

This dataset~\cite{imagenette} is composed of 3 color-channel images that are substantially larger than CIFAR-10 and CIFAR-100 with more complex features.  The raw images vary in size, but were all resized to 299\(\times{}\)299.  To cope with the larger image size, 15 convolutional layers were used with the first convolutional layer having a stride of 2, and given the addition of 16 feature maps per convolutional operation, this resulted in the presence of 256 feature maps after the final convolution.  Max pooling was applied after the fifth and tenth convolutional operations.  Using this configuration, the final set of feature maps were 10\(\times{}\)10, consistent with all other datasets experimented on.  Due to the larger number of parameters required for this network, a batch size of 32 was used, as was dictated by the constraints of available hardware.  Data augmentation used during the training for this dataset was the same strategy as used for the experiments in~\cite{Byerly2021a}.

\section{Baseline Results}\label{baseline_results}

All subsequent experiments detailed in this chapter involve excluding some subset of the training data during training.  In order to understand the impact of those exclusions, a set of baseline experiments was conducted for which all training samples were included for all of the investigated datasets.  The results of those experiments are presented in \autoref{tab:baseline_results}.  For these experiments and all subsequent experiments, five trials of each were conducted.

\begin{table}[!htbp]
    \caption{Baseline Results Wherein No Training Samples Were Excluded}\label{tab:baseline_results}
    \begin{tabularx}{\textwidth}{@{}Xrr@{}}
        \toprule
          Dataset & Accuracy & Standard Deviation \\
        \midrule
          MNIST         & 99.716\% & 0.000162481 \\
          Fashion-MNIST & 93.404\% & 0.001380724 \\
          CIFAR-10      & 89.146\% & 0.001518684 \\
          CIFAR-100     & 61.896\% & 0.001786169 \\
          Imagenette    & 92.390\% & 0.002333238 \\
        \bottomrule
    \end{tabularx}
\end{table}

\section{Data Reduction Strategies}\label{data_reduction_strategies}

\subsection{Experimental Design}\label{data_reduction_strategies_experimental_design}

\begin{figure}[!htbp]
  \centering
  \begin{minipage}[t]{.49\textwidth}
    \centering
    \includegraphics[width=\linewidth]{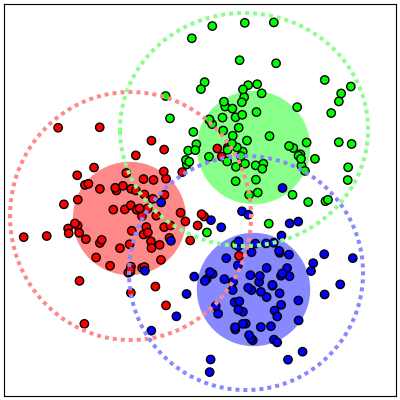}
    \caption{Visualization of 2-Dimensional Data Generated for 3 Relatively Well Separated Classes (All Points)}\label{fig:exclusion_2d_example_none}
  \end{minipage}
  \hfill
  \begin{minipage}[t]{.49\textwidth}
    \centering
    \includegraphics[width=\linewidth]{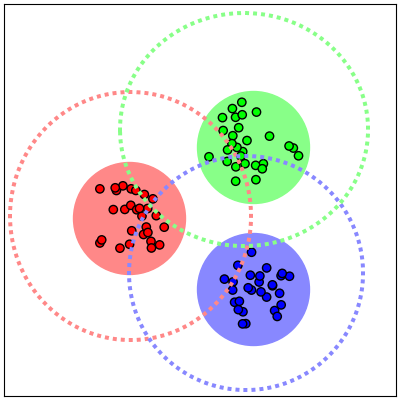}
    \caption{Visualization of 2-Dimensional Data Generated for 3 Relatively Well Separated Classes (\textit{Lateral Exclusion} Visualized)}\label{fig:exclusion_2d_example_lateral}
  \end{minipage}
\end{figure}

\begin{figure}[!htbp]
  \centering
  \begin{minipage}[t]{.49\textwidth}
    \centering
    \includegraphics[width=\linewidth]{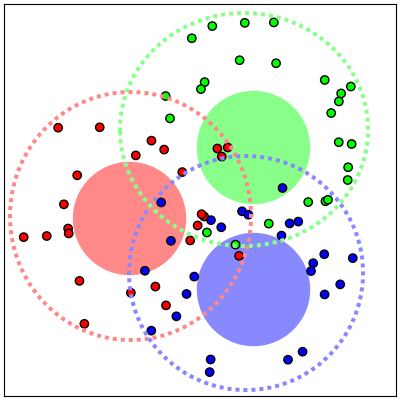}
    \caption{Visualization of 2-Dimensional Data Generated for 3 Relatively Well Separated Classes (\textit{Central Exclusion} Visualized)}\label{fig:exclusion_2d_example_central}
  \end{minipage}
  \hfill
  \begin{minipage}[t]{.49\textwidth}
    \centering
    \includegraphics[width=\linewidth]{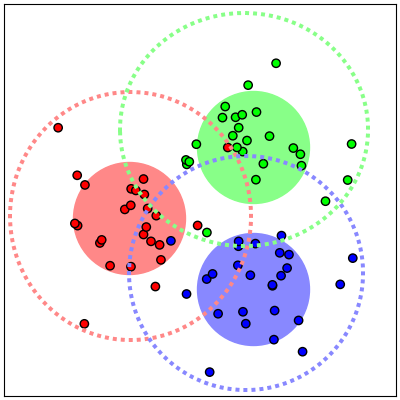}
    \caption{Visualization of 2-Dimensional Data Generated for 3 Relatively Well Separated Classes (\textit{Random Exclusion} Visualized)}\label{fig:exclusion_2d_example_random}
  \end{minipage}
\end{figure}

For the first set of data reduction experiments detailed in this work, the high-dimensional image data was reduced using UMAP to 3 dimensions.  Then, three data reduction strategies for selecting the data to exclude during training were considered.
For the first two data reduction strategies, the 3 dimensional centroid for each class was calculated.  The \textit{Lateral Exclusion} data reduction strategy excluded samples furthest from the centroid (compare \autoref{fig:exclusion_2d_example_none} to \autoref{fig:exclusion_2d_example_lateral}).  The \textit{Central Exclusion} data reduction strategy excluded samples nearest to the centroid (compare \autoref{fig:exclusion_2d_example_none} to \autoref{fig:exclusion_2d_example_central}).  The \textit{Random Exclusion} data reduction strategy excluded samples randomly (compare \autoref{fig:exclusion_2d_example_none} to \autoref{fig:exclusion_2d_example_random}).  Then for each data reduction strategy, experiments were conducted excluding 1\%, 2\%, 5\%, 10\%, 25\%, and 50\% of the training data.  These experiments were conducted on MNIST, Fashion-MNIST, CIFAR-10, CIFAR-100, and Imagenette.  

\subsection{Experimental Results}\label{data_reduction_strategies_experimental_results}

\autoref{tab:reduction_strategy_results_mnist} shows the results of the experiments for the MNIST dataset.  Using the \textit{Random Exclusion} strategy was inferior in all cases.  Using the \textit{Lateral Exclusion} strategy produced a mean accuracy of 0.002\% greater than when using the \textit{Central Exclusion} strategy when 2\% of the data was excluded, which is not a statistically significant difference.  For all other levels of exclusion, using the \textit{Central Exclusion} strategy was statically significantly superior.  When excluding 5\% of the training data using the \textit{Central Exclusion} strategy, the accuracy achieved was higher than the baseline that included all samples, though not enough trials were conducted to confirm this as statistically significant.  When excluding 10\%, the accuracy was identical to that of the baseline.  When using the \textit{Central Exclusion} strategy, only when excluding 25\% or 50\% was the accuracy statistically significantly lower than the baseline. 

\autoref{tab:reduction_strategy_results_fashion_mnist} shows the results of the experiments for the Fashion-MNIST dataset.  In all cases, the \textit{Central Exclusion} strategy proved statistically significantly superior to either of the other two strategies.  The \textit{Lateral Exclusion} strategy was superior to the \textit{Random Exclusion} strategy in all but the case of 2\% exclusion.  However, this superiority was only shown to be statistically significant in the case of the 25\% exclusion.  There was no statistically significant difference from the baseline when excluding either 1\% or 2\% of the training data.

The experiments performed on CIFAR-10 (see \autoref{tab:reduction_strategy_results_cifar10})
demonstrated the greatest amount of ambiguity among all datasets.  All three exclusion strategies outperformed the baseline for exclusions of both 1\% and 2\%, although not enough trials were conducted to show this to be statistically significant.  The \textit{Lateral Exclusion} strategy achieved the highest accuracy when excluding 1\%, the \textit{Random Exclusion} strategy achieved the highest accuracy when excluding 10\%, and the \textit{Central Exclusion} strategy achieved the highest accuracy in all other cases.  The statistically significant differences occurred when excluding 5\%, 25\%, and 50\%.  When excluding 5\%, the \textit{Central Exclusion} strategy was statistically significantly superior to both of the other two strategies.  When excluding 25\% and 50\%, the \textit{Central Exclusion} strategy was superior to both of the other two strategies, but only statistically significantly superior to the \textit{Lateral Exclusion} strategy.

\autoref{tab:reduction_strategy_results_cifar100} shows the results of the experiments for the CIFAR-100 dataset.  In all cases, the accuracy when using the \textit{Central Exclusion} was superior to the other two strategies.  The superiority was shown to be statistically significant in the cases of the 10\%, 25\%, and 50\% exclusion experiments.  There was no statistically significant difference from the baseline when excluding either 1\% or 2\% of the training data.

\autoref{tab:reduction_strategy_results_imagenette} shows the results of the experiments for the Imagenette dataset.  The \textit{Random Exclusion} strategy achieved a statistically insignificant superior accuracy relative to the other two methods for the 1\% and 2\% exclusion experiments.  For the remainder of the experiments, the \textit{Central Exclusion} strategy achieved a higher accuracy than the other two methods, although this was only statistically significant for 3 out of 4 such experiments (5\%, 25\%, and 50\% exclusion).  There was no statistically significant difference from the baseline when excluding 1\%  the training data.

After examining these five datasets, it can safely be concluded that, in general, the \textit{Central Exclusion} strategy is the superior strategy among the three.  In no experiments did either of the other two strategies show a statistically significant superiority to it.  Excluding data resulted in equivalent or superior accuracy over the baseline for experiments using 3 of the 5 datasets, including CIFAR-10 when excluding 1\%, 2\%, or 5\%, CIFAR-100 when excluding 1\%, and MNIST when excluding 5\% or 10\%.  However, due to the high variance across trials, this was not able to be demonstrated as statistically significant.  Similarly, no experiment for any dataset when excluding 1\% or 2\% was shown to be statistically significantly inferior to the baseline.

\begin{table}[!htbp]
  \caption{Data Reduction Strategy Experimental Results --- MNIST}\label{tab:reduction_strategy_results_mnist}
  \begin{tabularx}{\textwidth}{@{}Lrrrrrrr@{}}
      \toprule
        & \multicolumn{2}{r}{\underline{Lateral Exclusion}}
        & \multicolumn{2}{r}{\underline{Central Exclusion}}
        & \multicolumn{2}{r}{\underline{Random Exclusion}} \\
        Excl. \% &  Acc. & Std. Dev. &  Acc. & Std. Dev. & Acc. & Std. Dev. \\
      \midrule
           1\% & 99.702\% & \num{1.47e-4} & \textbf{99.714\%} & \num{1.74e-4} & 99.694\% & \num{4.90e-5} \\
           2\% & \textbf{99.714\%} & \num{2.42e-4} & 99.712\% & \num{1.17e-4} & 99.696\% & \num{2.33e-4} \\
           5\% & 99.676\% & \num{1.62e-4} & \textbf{99.720\%} & \num{1.67e-4} & 99.658\% & \num{1.83e-4} \\
          10\% & 99.658\% & \num{1.17e-4} & \textbf{99.716\%} & \num{1.20e-4} & 99.638\% & \num{2.32e-4} \\
          25\% & 99.546\% & \num{2.58e-4} & \textbf{99.698\%} & \num{2.14e-4} & 99.526\% & \num{2.33e-4} \\
          50\% & 99.190\% & \num{3.03e-4} & \textbf{99.686\%} & \num{2.65e-4} & 99.236\% & \num{2.58e-4} \\
      \bottomrule
  \end{tabularx}\\[0.025in]
  \captionsetup{justification=raggedright,singlelinecheck=false}
  \caption*{The baseline accuracy, excluding no samples, for this dataset was 99.716\%}
\end{table}

\begin{table}[!htbp]
    \caption{Data Reduction Strategy Experimental Results --- Fashion-MNIST}\label{tab:reduction_strategy_results_fashion_mnist}
    \begin{tabularx}{\textwidth}{@{}Lrrrrrrr@{}}
      \toprule
        & \multicolumn{2}{r}{\underline{Lateral Exclusion}}
        & \multicolumn{2}{r}{\underline{Central Exclusion}}
        & \multicolumn{2}{r}{\underline{Random Exclusion}} \\
        Excl. \% &  Acc. & Std. Dev. &  Acc. & Std. Dev. & Acc. & Std. Dev. \\
      \midrule
           1\% & 93.114\% & \num{7.89e-4} & \textbf{93.346\%} & \num{8.71e-4} & 93.032\% & \num{1.33e-3} \\
           2\% & 92.896\% & \num{6.56e-4} & \textbf{93.324\%} & \num{1.75e-3} & 92.940\% & \num{1.66e-3} \\
           5\% & 92.096\% & \num{1.33e-3} & \textbf{93.198\%} & \num{5.49e-4} & 92.084\% & \num{7.42e-4} \\
          10\% & 91.088\% & \num{1.34e-3} & \textbf{93.184\%} & \num{6.47e-4} & 91.032\% & \num{4.71e-4} \\
          25\% & 89.096\% & \num{9.00e-4} & \textbf{92.162\%} & \num{9.45e-4} & 88.932\% & \num{5.38e-4} \\
          50\% & 85.690\% & \num{4.24e-4} & \textbf{88.692\%} & \num{2.31e-3} & 85.558\% & \num{1.71e-3} \\
        \bottomrule
    \end{tabularx}\\[0.025in]
    \captionsetup{justification=raggedright,singlelinecheck=false}
    \caption*{The baseline accuracy, excluding no samples, for this dataset was 93.404\%}
\end{table}

\begin{table}[!htbp]
    \caption{Data Reduction Strategy Experimental Results --- CIFAR-10}\label{tab:reduction_strategy_results_cifar10}
    \begin{tabularx}{\textwidth}{@{}Lrrrrrrr@{}}
      \toprule
        & \multicolumn{2}{r}{\underline{Lateral Exclusion}}
        & \multicolumn{2}{r}{\underline{Central Exclusion}}
        & \multicolumn{2}{r}{\underline{Random Exclusion}} \\
        Excl. \% &  Acc. & Std. Dev. &  Acc. & Std. Dev. & Acc. & Std. Dev. \\
      \midrule
           1\% & \textbf{89.332\%} & \num{2.83e-3} & 89.280\% & \num{1.86e-3} & 89.318\% & \num{1.18e-3} \\
           2\% & 89.148\% & \num{8.93e-4} & \textbf{89.206\%} & \num{1.21e-3} & 89.148\% & \num{1.04e-3} \\
           5\% & 89.054\% & \num{1.45e-3} & \textbf{89.224\%} & \num{1.30e-3} & 89.086\% & \num{8.09e-4} \\
          10\% & 88.708\% & \num{6.97e-4} & 88.620\% & \num{1.87e-3} & \textbf{88.838\%} & \num{1.94e-3} \\
          25\% & 87.098\% & \num{1.44e-3} & \textbf{87.930\%} & \num{4.00e-4} & 87.878\% & \num{4.29e-3} \\
          50\% & 84.044\% & \num{1.32e-3} & \textbf{85.900\%} & \num{1.13e-3} & 85.706\% & \num{7.40e-3} \\
        \bottomrule
    \end{tabularx}\\[0.025in]
    \captionsetup{justification=raggedright,singlelinecheck=false}
    \caption*{The baseline accuracy, excluding no samples, for this dataset was 89.146\%}
\end{table}

\begin{table}[!htbp]
    \caption{Data Reduction Strategy Experimental Results --- CIFAR-100}\label{tab:reduction_strategy_results_cifar100}
    \begin{tabularx}{\textwidth}{@{}Lrrrrrrr@{}}
      \toprule
        & \multicolumn{2}{r}{\underline{Lateral Exclusion}}
        & \multicolumn{2}{r}{\underline{Central Exclusion}}
        & \multicolumn{2}{r}{\underline{Random Exclusion}} \\
        Excl. \% &  Acc. & Std. Dev. &  Acc. & Std. Dev. & Acc. & Std. Dev. \\
      \midrule
           1\% & 61.926\% & \num{2.37e-3} & \textbf{61.940\%} & \num{3.72e-3} & 61.888\% & \num{4.26e-3} \\
           2\% & 61.766\% & \num{2.63e-3} & \textbf{61.866\%} & \num{1.90e-3} & 61.594\% & \num{2.44e-3} \\
           5\% & 61.262\% & \num{3.26e-3} & \textbf{61.414\%} & \num{2.07e-3} & 61.102\% & \num{3.81e-3} \\
          10\% & 60.410\% & \num{2.72e-3} & \textbf{61.044\%} & \num{2.53e-3} & 60.012\% & \num{2.41e-3} \\
          25\% & 57.440\% & \num{2.25e-3} & \textbf{59.158\%} & \num{2.01e-3} & 57.206\% & \num{4.07e-3} \\
          50\% & 50.754\% & \num{2.48e-3} & \textbf{53.868\%} & \num{4.45e-3} & 51.334\% & \num{3.94e-3} \\
        \bottomrule
    \end{tabularx}\\[0.025in]
    \captionsetup{justification=raggedright,singlelinecheck=false}
    \caption*{The baseline accuracy, excluding no samples, for this dataset was 61.896\%}
\end{table}

\begin{table}[!htbp]
    \caption{Data Reduction Strategy Experimental Results --- Imagenette}\label{tab:reduction_strategy_results_imagenette}
    \begin{tabularx}{\textwidth}{@{}Lrrrrrrr@{}}
      \toprule
        & \multicolumn{2}{r}{\underline{Lateral Exclusion}}
        & \multicolumn{2}{r}{\underline{Central Exclusion}}
        & \multicolumn{2}{r}{\underline{Random Exclusion}} \\
        Excl. \% &  Acc. & Std. Dev. &  Acc. & Std. Dev. & Acc. & Std. Dev. \\
      \midrule
           1\% & 92.288\% & \num{1.46e-3} & 92.300\% & \num{3.71e-3} & \textbf{92.316\%} & \num{2.34e-3} \\
           2\% & 92.022\% & \num{2.14e-3} & 92.196\% & \num{2.28e-3} & \textbf{92.342\%} & \num{2.19e-3} \\
           5\% & 91.896\% & \num{6.77e-4} & \textbf{92.214\%} & \num{1.57e-3} & 91.994\% & \num{8.36e-4} \\
          10\% & 91.544\% & \num{2.91e-3} & \textbf{91.844\%} & \num{2.91e-3} & 91.680\% & \num{1.94e-3} \\
          25\% & 90.422\% & \num{2.20e-3} & \textbf{90.998\%} & \num{2.35e-3} & 90.342\% & \num{1.78e-3} \\
          50\% & 87.998\% & \num{3.55e-3} & \textbf{88.174\%} & \num{2.56e-3} & 87.914\% & \num{5.66e-3} \\
        \bottomrule
    \end{tabularx}\\[0.025in]
    \captionsetup{justification=raggedright,singlelinecheck=false}
    \caption*{The baseline accuracy, excluding no samples, for this dataset was 92.390\%}
\end{table}

\section{Dimensions to Reduce to}\label{dimensions_to_reduce_to}

\subsection{Experimental Design}\label{dimensions_to_reduce_to_experimental_design}

To test the hypothesis that 3-dimensions was the appropriate choice for the dimensional reduction, an additional set of experiments using the \textit{Central Exclusion} data reduction method, and excluding 1\%, 2\%, 5\%, 10\%, 25\%, and 50\% of the training data were conducted.  This set of experiments used 2, 5, and 10 dimensional reductions and were conducted on CIFAR-10, CIFAR-100, MNIST, Fashion-MNIST, and Imagenette.

\subsection{Experimental Results}\label{dimensions_to_reduce_to_experimental_results}

Executing 5 trials of each of 6 different amounts of excluded data results in 30 total trials per dataset and number of dimensions being used to determine the exclusion.  For each of the 2, 5, and 10 dimensional reductions we compared the mean accuracy achieved across all 30 trials to the 30 trials of the experiments that used a 3 dimensional reduction.  \autoref{tab:dimensions_to_reduce_to_analysis} shows the results of those comparisons.  The comparisons showed no statistically significant difference between any paired sets of 30 trials.  5 out of 15 experiments showed a statistically insignificant superiority when using a 3 dimensional reduction, including all 3 comparisons of MNIST and 2 out of 3 comparisons of Imagenette.  Although, not reaching a reasonable threshold for statistical significance (\(p < 0.05\)), the MNIST comparisons had the lowest p-values. At first, this may seem surprising, but upon reflection, it seems reasonable that points UMAP places far from a class's centroid in 3 dimensions would also be likely to be placed far from the class's centroid in 2, 5, or 10 dimensions as well.

\begin{table}[!htbp]
  \caption{Comparison of Mean Accuracies for Exclusions Based on Differing Dimensional Reductions}\label{tab:dimensions_to_reduce_to_analysis}
  \begin{tabularx}{\textwidth}{@{}Xcrrr@{}}
      \toprule
        Dataset & Dimensions & Accuracy & 3D Accuracy & p-value \\
      \midrule
        MNIST         &  2 & 99.7000\% & \textbf{99.7077\%} & 0.111615712 \\
        MNIST         &  5 & 99.7027\% & \textbf{99.7077\%} & 0.232822313 \\
        MNIST         & 10 & 99.6977\% & \textbf{99.7077\%} & 0.083150845 \\
      \midrule
        Fashion-MNIST &  2 & \textbf{92.3803\%} & 92.3177\% & 0.44224238  \\
        Fashion-MNIST &  5 & \textbf{92.4217\%} & 92.3177\% & 0.405938742 \\
        Fashion-MNIST & 10 & \textbf{92.3490\%} & 92.3177\% & 0.471922235 \\
      \midrule
        CIFAR-10      &  2 & \textbf{88.3677\%} & 88.3600\% & 0.490392657 \\
        CIFAR-10      &  5 & \textbf{88.3710\%} & 88.3600\% & 0.486238005 \\
        CIFAR-10      & 10 & \textbf{88.3747\%} & 88.3600\% & 0.481346744 \\
      \midrule
        CIFAR-100     &  2 & \textbf{59.9347\%} & 59.8817\% & 0.471617385 \\
        CIFAR-100     &  5 & \textbf{60.0760\%} & 59.8817\% & 0.396706608 \\
        CIFAR-100     & 10 & \textbf{59.9790\%} & 59.8817\% & 0.448318135 \\
      \midrule
        Imagenette    &  2 & \textbf{91.3237\%} & 91.2877\% & 0.463151304 \\
        Imagenette    &  5 & 91.2153\% & \textbf{91.2877\%} & 0.427306126 \\
        Imagenette    & 10 & 91.2417\% & \textbf{91.2877\%} & 0.453380991 \\
      \bottomrule
  \end{tabularx}
\end{table}

\section{Distributions of the Dimensional Reductions}\label{distributions_of_the_dimensional_reductions}

\begin{figure}[!htbp]
  \centering
  \begin{minipage}[t]{.49\textwidth}
    \centering
    \includegraphics[width=\linewidth]{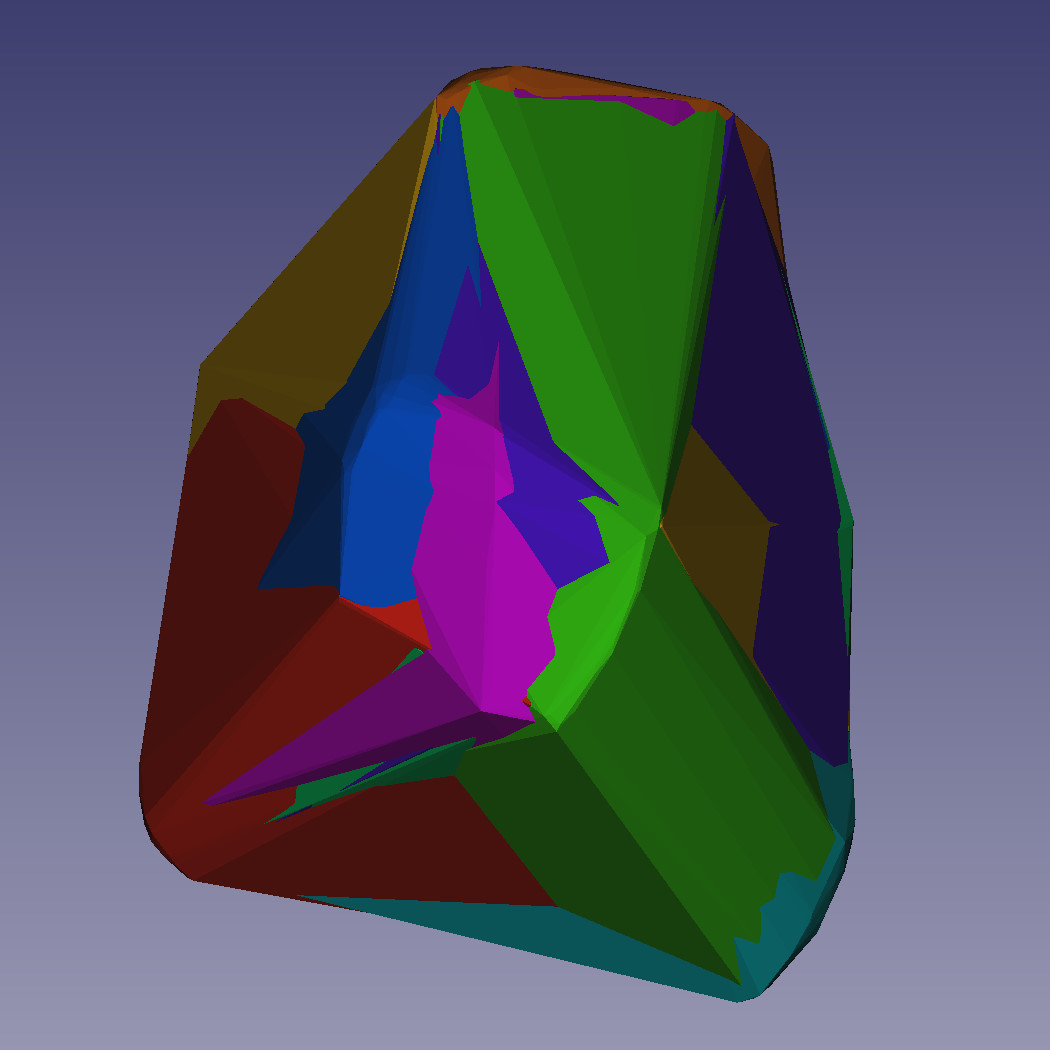}
    \caption{Visualization of 3-Dimensional Reduction of MNIST}\label{fig:mnist_3d_all}
  \end{minipage}
  \hfill
  \begin{minipage}[t]{.49\textwidth}
    \centering
    \includegraphics[width=\linewidth]{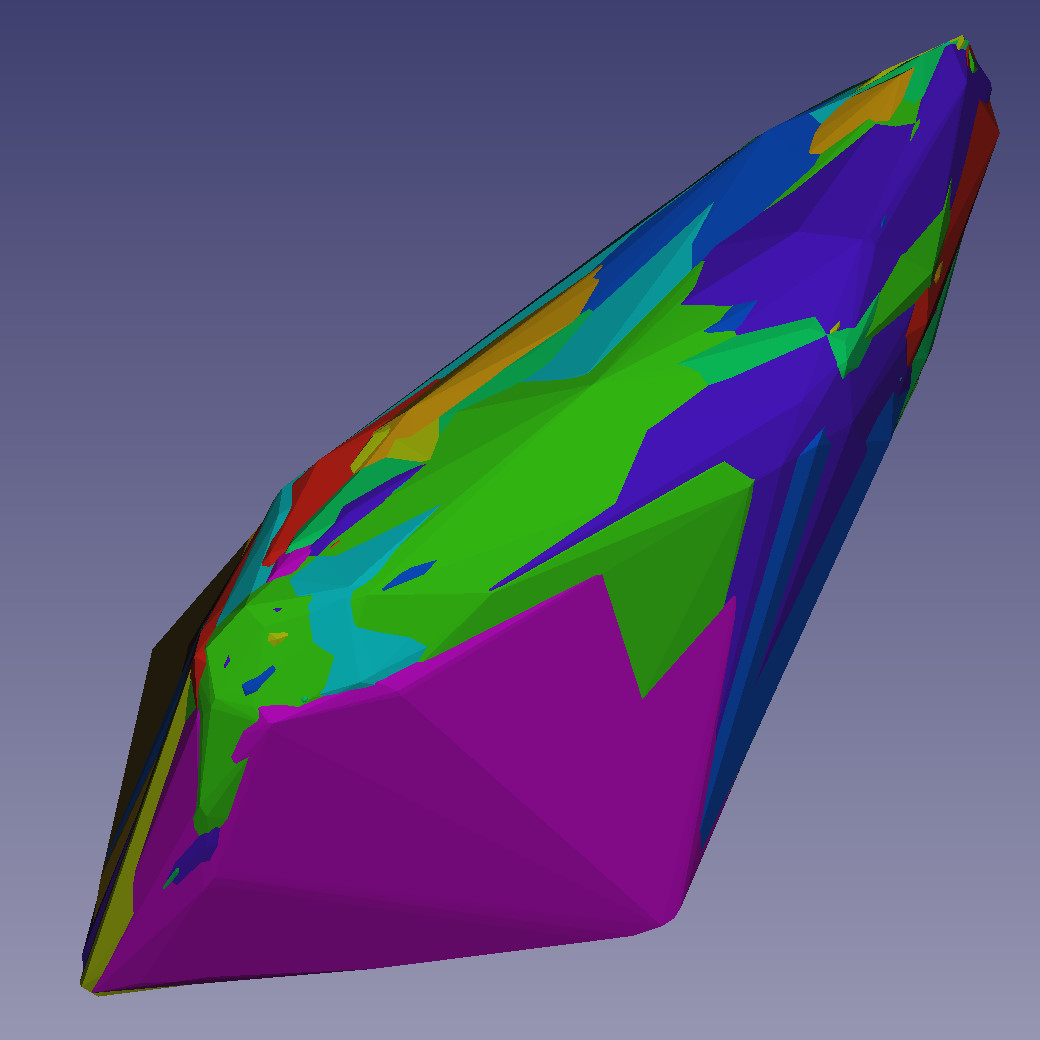}
    \caption{Visualization of 3-Dimensional Reduction of Imagenette}\label{fig:imagenette_3d_all}
  \end{minipage}
\end{figure}

UMAP generates numeric values for each dimension of the reduction performed.  Visualizations of 2-dimensional and 3-dimensional reductions are usually generated by drawing these points as small epsilon balls around the positions of those numeric values.  While these visualizations can provide some sense of where the classes' data are located in space, drawing convex hulls that surround each class's points provides a sharper distinction between the boundaries of the classes in the space.  \autoref{fig:mnist_3d_all} and \autoref{fig:imagenette_3d_all} show the result of drawing these hulls around the 3-dimensional reductions of the full MNIST and Imagenette training data, respectively.  These visualizations make it clear that when including all of the training data, there is very little distinction of boundaries in the space.

However, when excluding some of the data furthest from each class's centroid, the classes' hulls start to separate, in some cases, partially and in some cases entirely.  The amount of data that must be elided to achieve this is referred to in this work as \textit{Dataset Severability}.

\begin{definition}[Dataset Severability]
  \textit{Dataset Severability} is a qualitative judgment regarding the number of outliers that must be removed from each class in an \(m\)-dimensional reduction of the dataset so that structure and/or clustering is able to be observed.
\end{definition}

\begin{figure}[!htbp]
    \centering
    \begin{minipage}[t]{.49\textwidth}
      \centering
      \includegraphics[width=\linewidth]{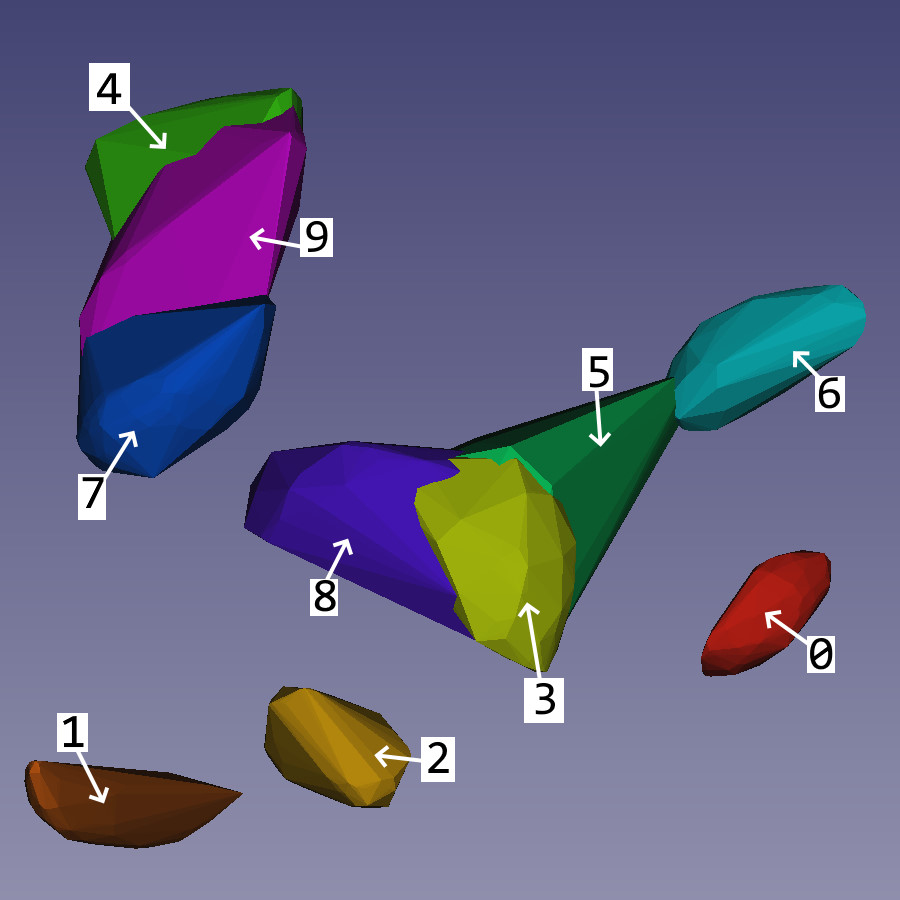}
      \caption{Visualization of 3-Dimensional Reduction of MNIST w/ Classes Labeled (Excluding Outliers)}\label{fig:mnist_3d}
    \end{minipage}
    \hfill
    \begin{minipage}[t]{.49\textwidth}
      \centering
      \includegraphics[width=\linewidth]{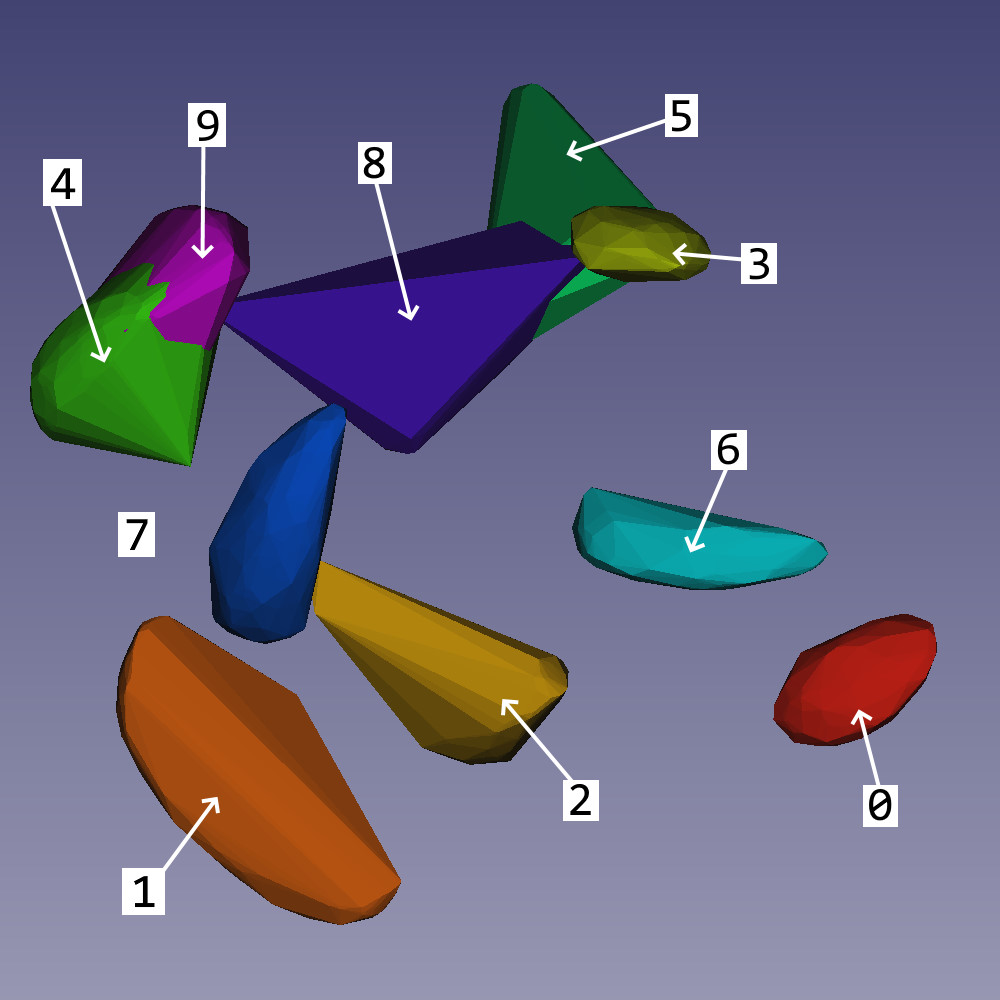}
      \caption{Visualization of 3-Dimensional Reduction of EMNIST-Digits w/ Classes Labeled (Excluding Outliers)}\label{fig:emnist_3d}
    \end{minipage}
\end{figure}

\autoref{fig:mnist_3d} and \autoref{fig:emnist_3d} show well severed classes for visualizations of the MNIST and EMNIST-Digits training data, respectively.  In each case, the data within three standard deviations of all dimensions was included, and data outside of these bounds was omitted.  This means that high severability is achieved by excluding \(\approx{}\) 0.27\% of the outliers for these datasets.  Despite consisting of images with the same size and number of color channels (monochromatic) as MNIST and EMNIST-Digits, Fashion-MNIST shows a dissimilar level of severability until all but one standard deviation from the centroid has been excluded (excluding \(\approx{}\) 31.73\%).

Especially interesting in the case of the Fashion-MNIST reduction is that there is a readily identifiable semantic difference between each of the four completely severed sets of overlapping hulls.  The set of ``overlapping'' hulls consisting of the single class ``trouser'' contains the only class in the dataset that is legwear.  The set of ``overlapping'' hulls consisting of the single class ``bag'' contains the only class in the dataset that is not a type of clothing that covers any part of the body.  The set of overlapping hulls consisting of the classes ``sneaker'', ``sandal'', and ``ankle boot'' contains the only classes in the dataset that are footwear.  Finally, the set of overlapping hulls consisting of the classes ``pullover'', ``coat'', ``shirt'', ``dress'', and ``t-shirt'' contains the only classes in the dataset that primarily cover the torso.

\begin{figure}[!htbp]
    \centering
    \begin{minipage}[t]{.49\textwidth}
      \centering
      \includegraphics[width=\linewidth]{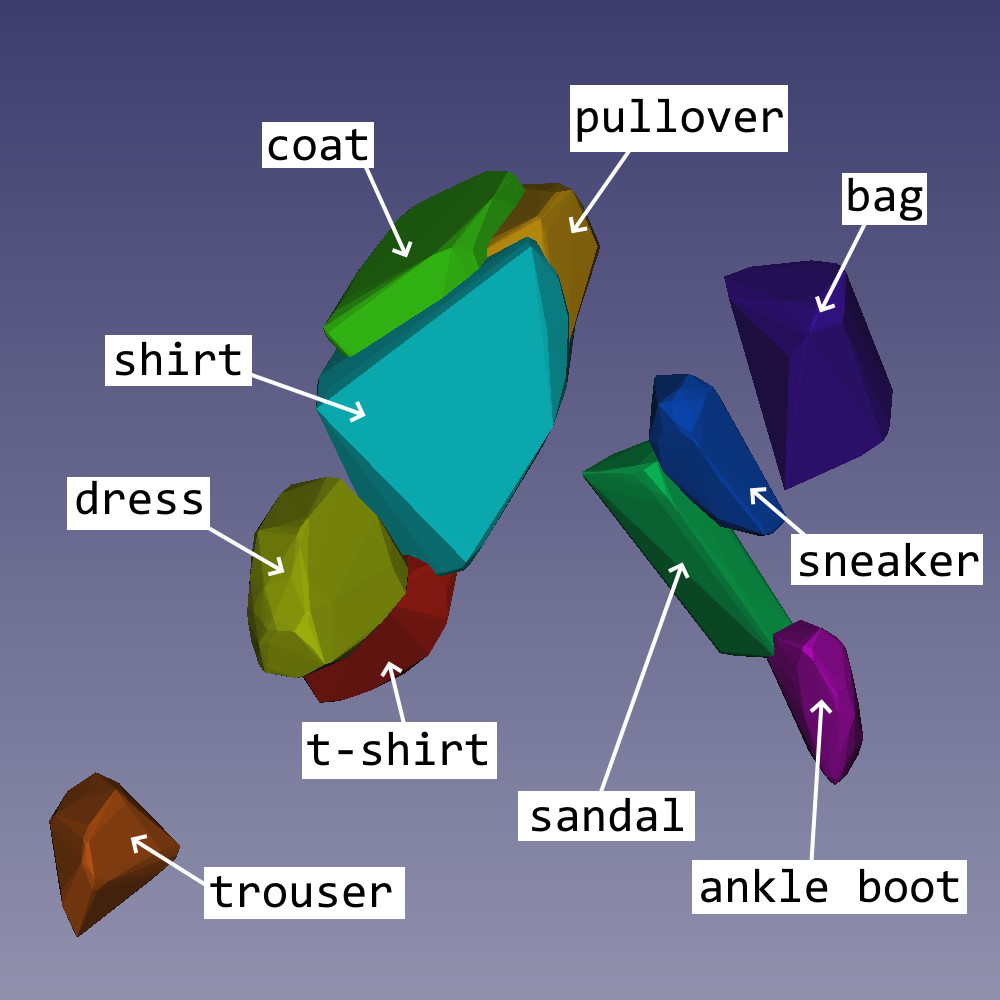}
      \caption{Visualization of 3-Dimensional Reduction of Fashion-MNIST w/ Classes Labeled (Excluding Outliers)}\label{fig:fashion_mnist_3d}
    \end{minipage}
    \hfill
    \begin{minipage}[t]{.49\textwidth}
      \centering
      \includegraphics[width=\linewidth]{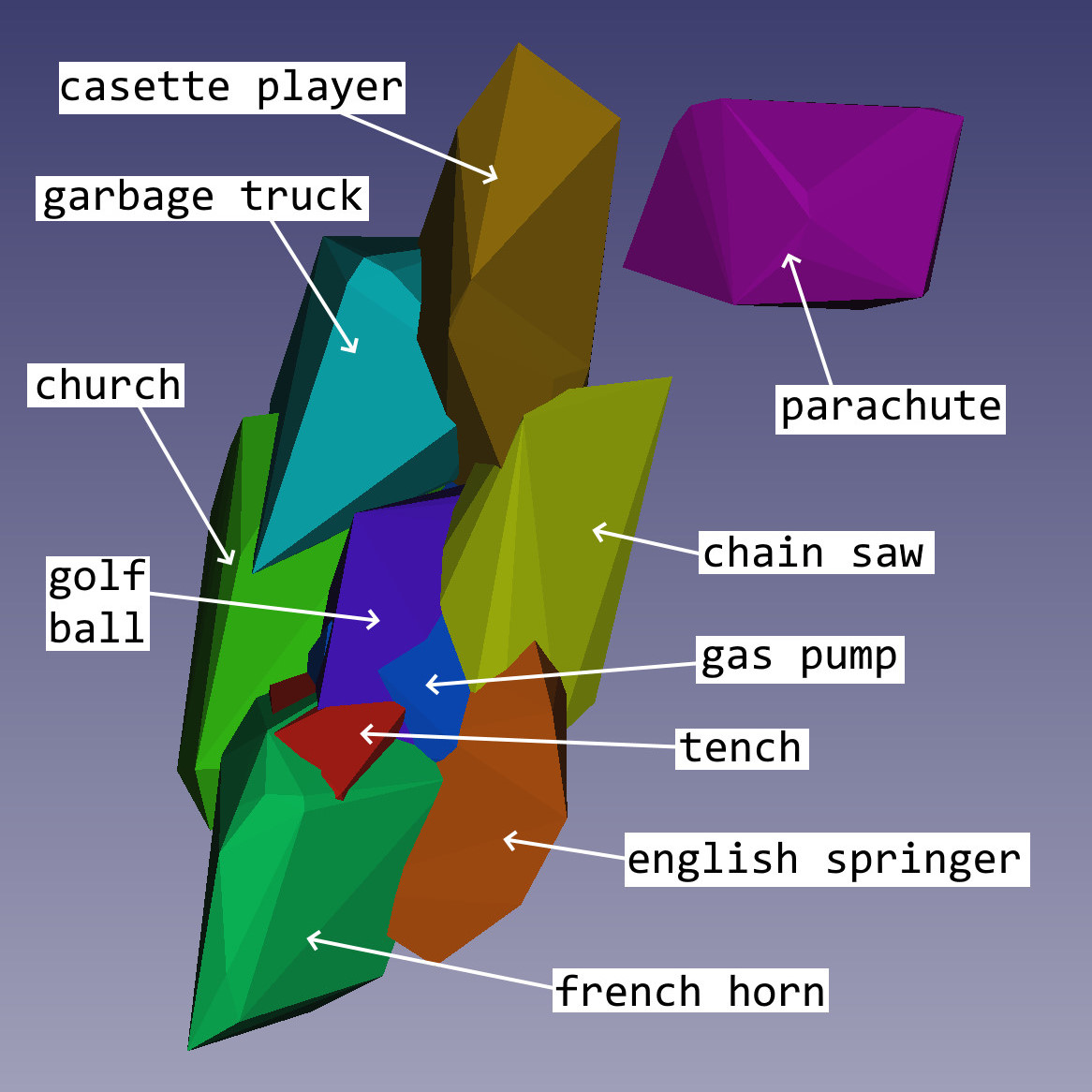}
      \caption{Visualization of 3-Dimensional Reduction of Imagenette w/ Classes Labeled (Excluding Outliers)}\label{fig:imagenette_3d}
    \end{minipage}
\end{figure}

\begin{figure}[!htbp]
    \centering
    \begin{minipage}[t]{.49\textwidth}
      \centering
      \includegraphics[width=\linewidth]{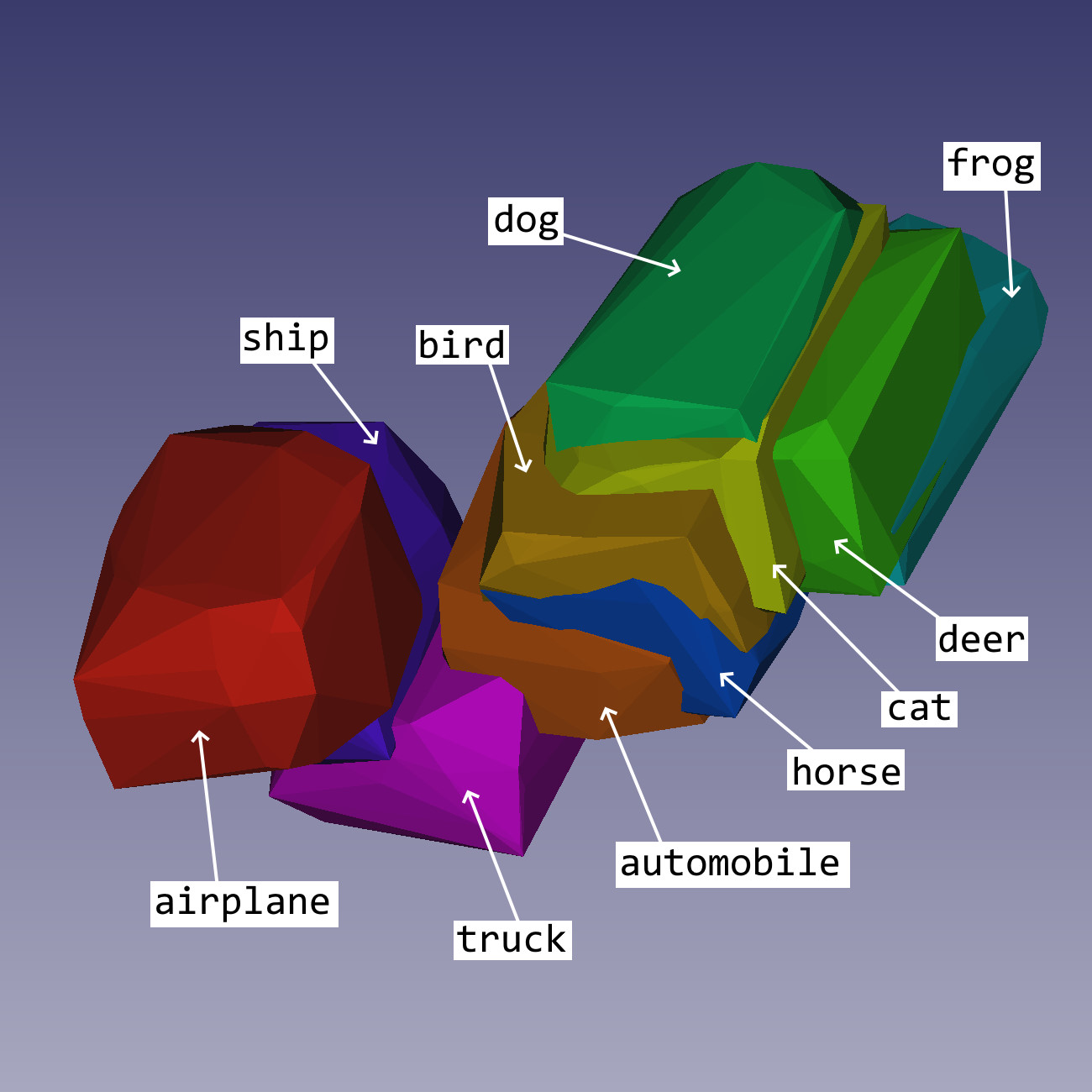}
      \caption{Visualization of 3-Dimensional Reduction of CIFAR-10 w/ Classes Labeled (Excluding Outliers)}\label{fig:cifar10_3d}
    \end{minipage}
    \hfill
    \begin{minipage}[t]{.49\textwidth}
      \centering
      \includegraphics[width=\linewidth]{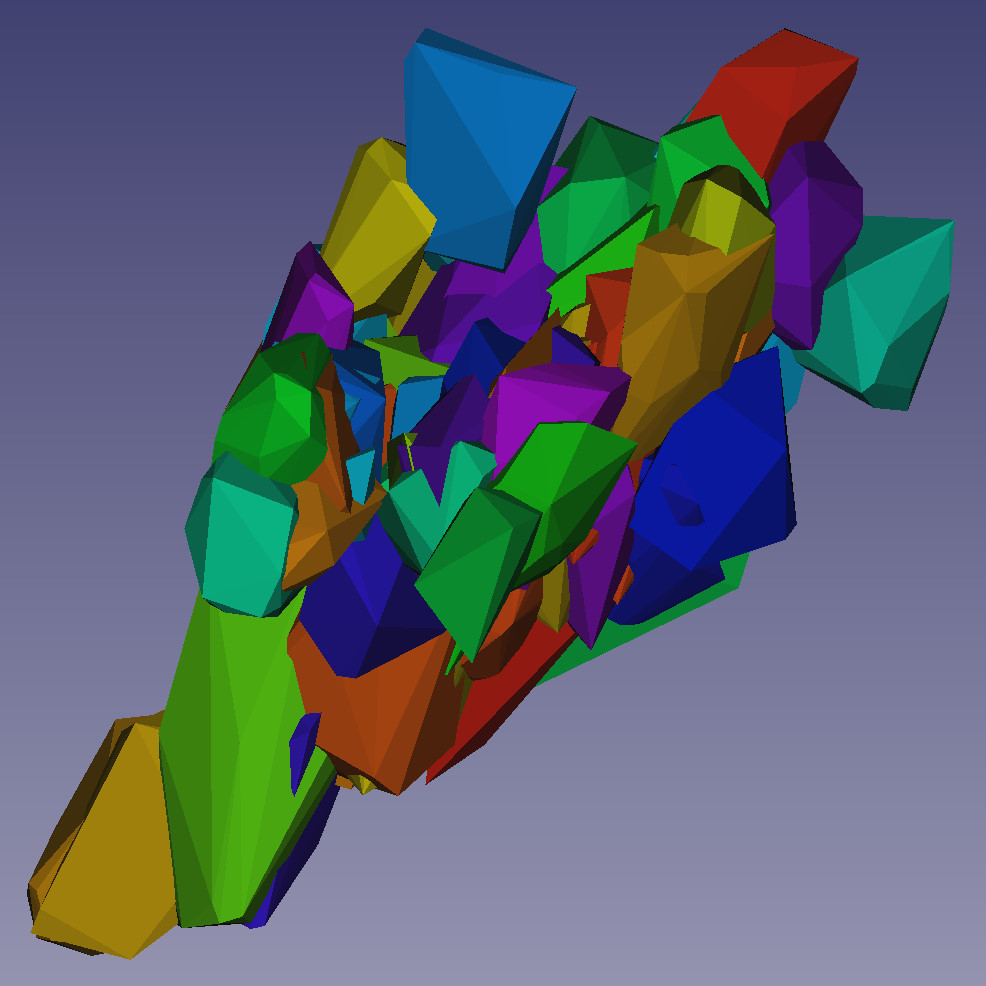}
      \caption{Visualization of 3-Dimensional Reduction of CIFAR-100 w/ Classes Labeled (Excluding Outliers)}\label{fig:cifar100_3d}
    \end{minipage}
\end{figure}

Imagenette (\autoref{fig:imagenette_3d}), CIFAR-10 (\autoref{fig:cifar10_3d}), and CIFAR-100 (\autoref{fig:cifar100_3d}) require all but those samples within half a standard deviation of each class's centroid to be elided before structure emerges (excluding \(\approx{}\) 61.71\%).  Of these three, the reduction of CIFAR-10 displays the most interesting semantic relationships among the classes' hulls locations in space.  The classes on the left of the visualization in \autoref{fig:cifar10_3d} are all man-made (specifically vehicles) whereas the classes on the right are all lifeforms.  Especially interesting among the lifeforms is that all of the mammals are grouped together with the one amphibian (frog) on the outer edge of the group. 

\begin{figure}[!htbp]
    \centering
    \begin{minipage}[t]{.49\textwidth}
      \centering
      \includegraphics[width=\linewidth]{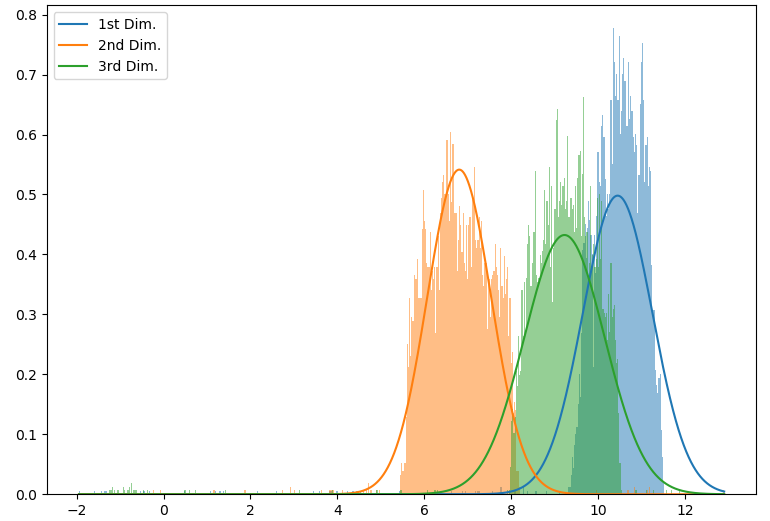}
      \caption{MNIST Reduction Distributions --- Class `0'}\label{fig:mnist_distribution_0}
    \end{minipage}
    \hfill
    \begin{minipage}[t]{.49\textwidth}
      \centering
      \includegraphics[width=\linewidth]{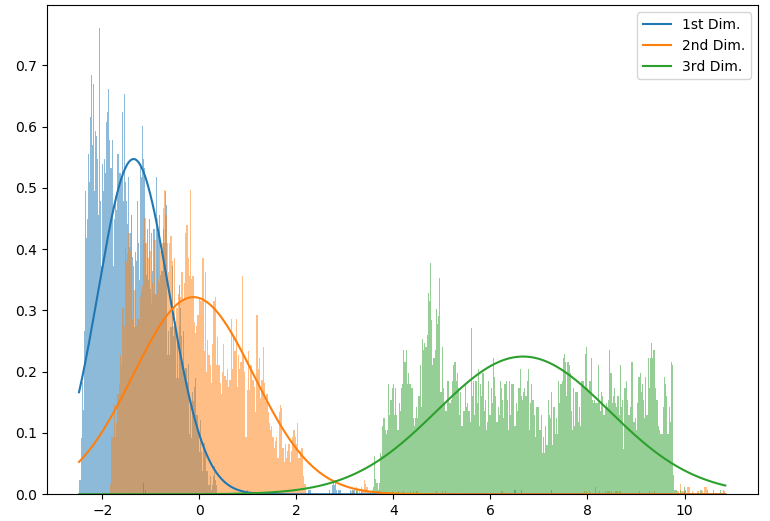}
      \caption{MNIST Reduction Distributions --- Class `1'}\label{fig:mnist_distribution_1}
    \end{minipage}
\end{figure}

\begin{figure}[!htbp]
    \centering
    \begin{minipage}[t]{.49\textwidth}
      \centering
      \includegraphics[width=\linewidth]{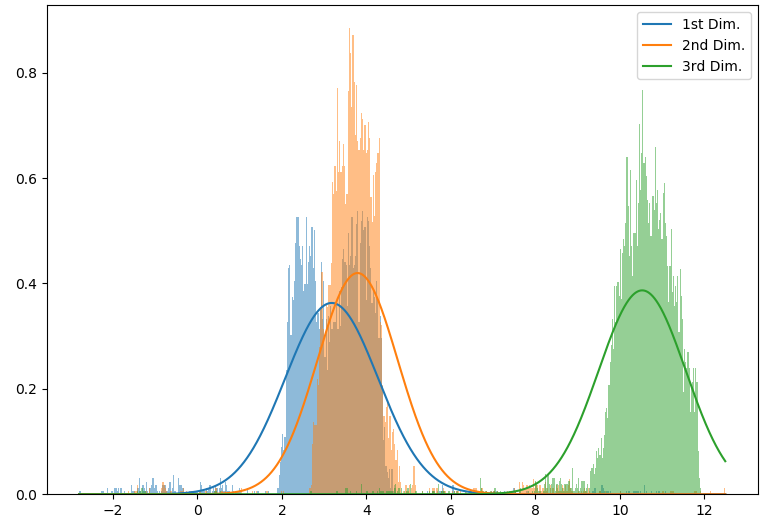}
      \caption{MNIST Reduction Distributions --- Class `2'}\label{fig:mnist_distribution_2}
    \end{minipage}
    \hfill
    \begin{minipage}[t]{.49\textwidth}
      \centering
      \includegraphics[width=\linewidth]{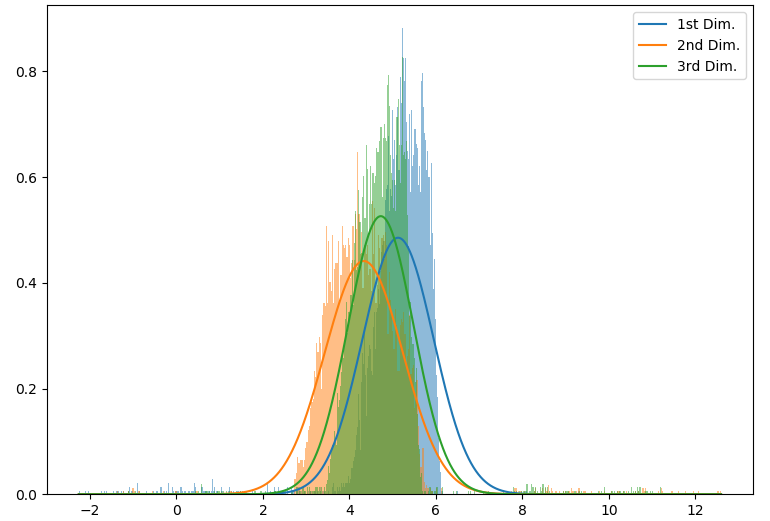}
      \caption{MNIST Reduction Distributions --- Class `3'}\label{fig:mnist_distribution_3}
    \end{minipage}
\end{figure}

\begin{figure}[!htbp]
    \centering
    \begin{minipage}[t]{.49\textwidth}
      \centering
      \includegraphics[width=\linewidth]{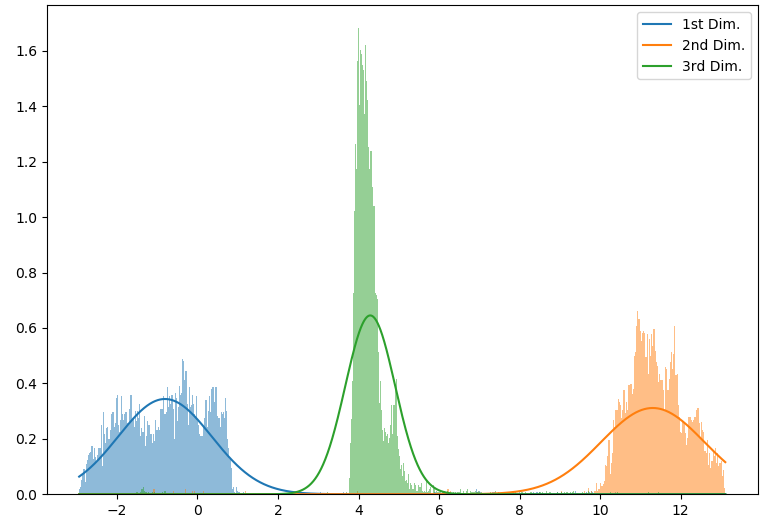}
      \caption{MNIST Reduction Distributions --- Class `4'}\label{fig:mnist_distribution_4}
    \end{minipage}
    \hfill
    \begin{minipage}[t]{.49\textwidth}
      \centering
      \includegraphics[width=\linewidth]{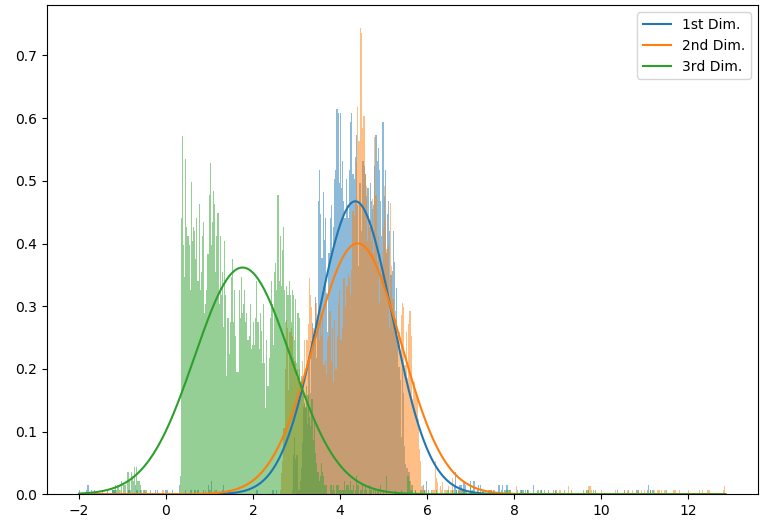}
      \caption{MNIST Reduction Distributions --- Class `5'}\label{fig:mnist_distribution_5}
    \end{minipage}
\end{figure}

\begin{figure}[!htbp]
    \centering
    \begin{minipage}[t]{.49\textwidth}
      \centering
      \includegraphics[width=\linewidth]{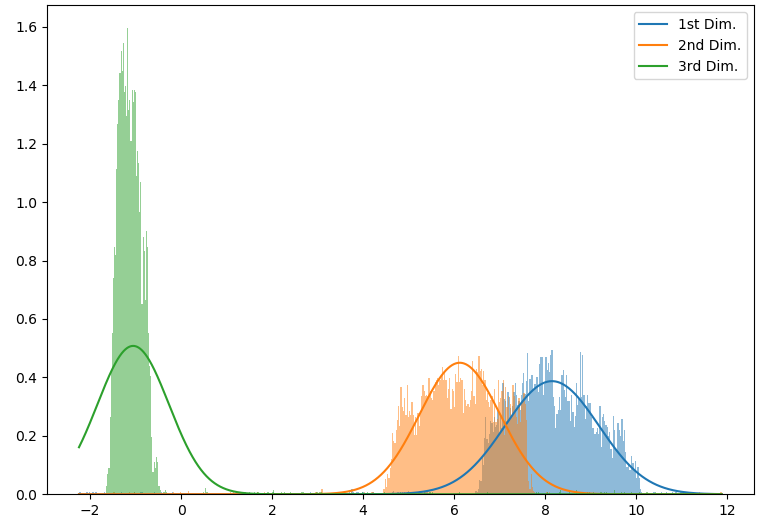}
      \caption{MNIST Reduction Distributions --- Class `6'}\label{fig:mnist_distribution_6}
    \end{minipage}
    \hfill
    \begin{minipage}[t]{.49\textwidth}
      \centering
      \includegraphics[width=\linewidth]{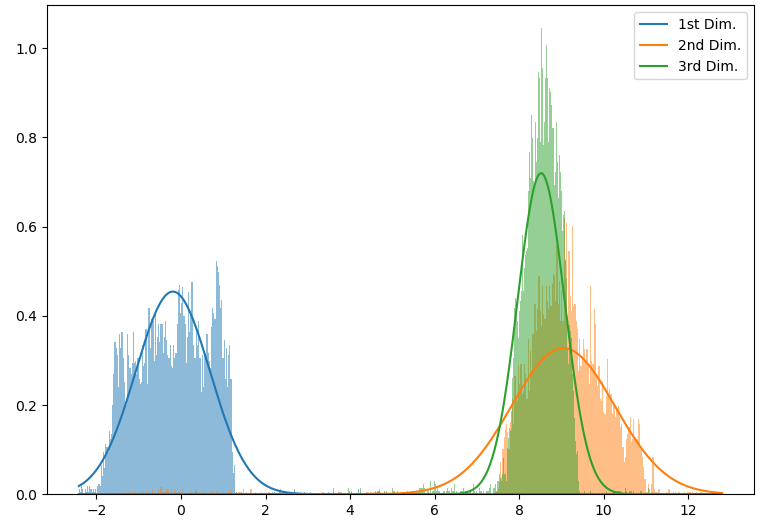}
      \caption{MNIST Reduction Distributions --- Class `7'}\label{fig:mnist_distribution_7}
    \end{minipage}
\end{figure}

\begin{figure}[!htbp]
    \centering
    \begin{minipage}[t]{.49\textwidth}
      \centering
      \includegraphics[width=\linewidth]{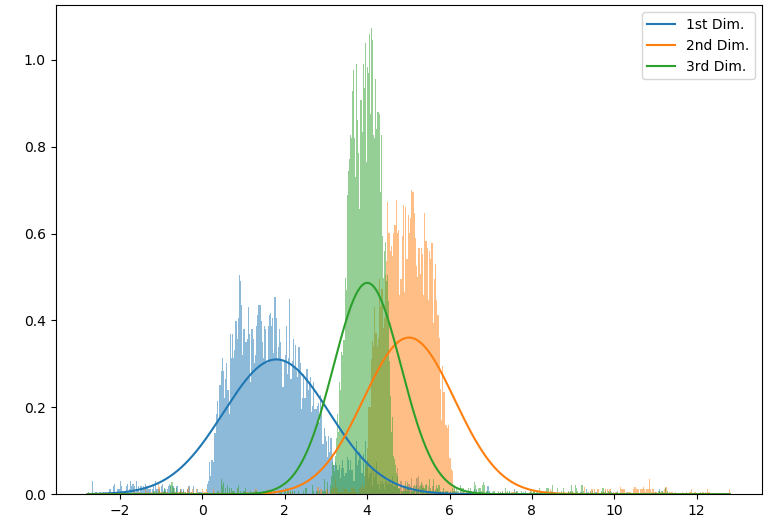}
      \caption{MNIST Reduction Distributions --- Class `8'}\label{fig:mnist_distribution_8}
    \end{minipage}
    \hfill
    \begin{minipage}[t]{.49\textwidth}
      \centering
      \includegraphics[width=\linewidth]{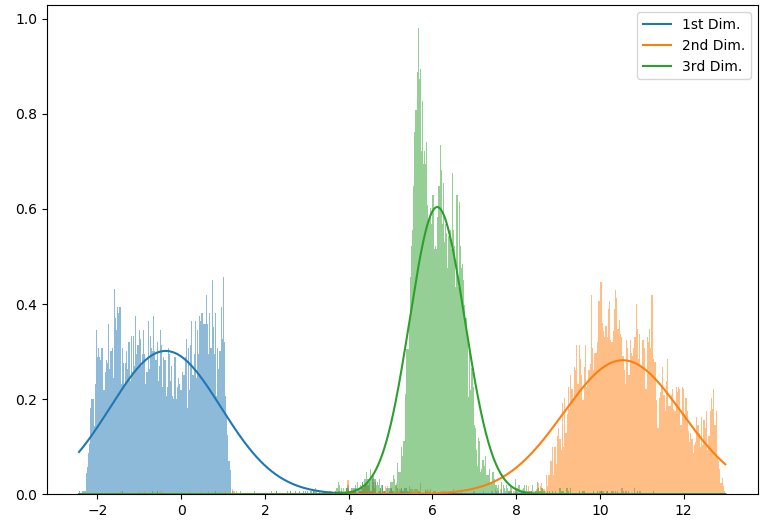}
      \caption{MNIST Reduction Distributions --- Class `9'}\label{fig:mnist_distribution_9}
    \end{minipage}
\end{figure}

\begin{figure}[!htbp]
    \centering
    \begin{minipage}[t]{.49\textwidth}
      \centering
      \includegraphics[width=\linewidth]{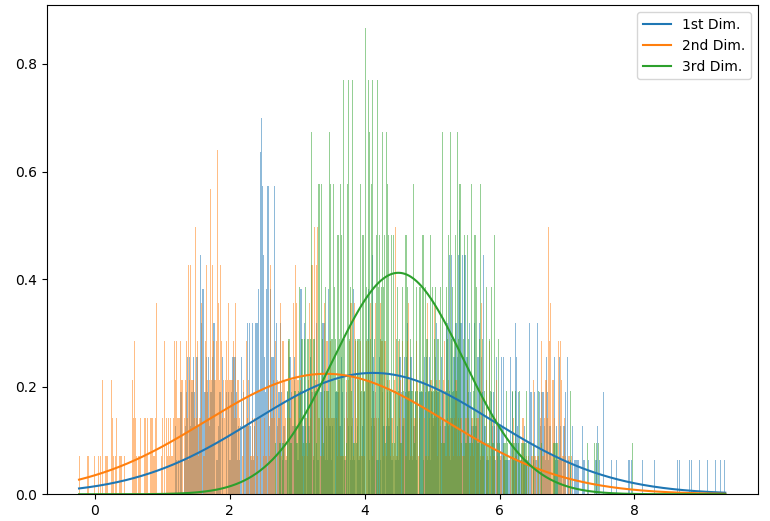}
      \caption{Imagenette Reduction Distributions --- Class `Tench'}\label{fig:imagenette_distribution_0}
    \end{minipage}
    \hfill
    \begin{minipage}[t]{.49\textwidth}
      \centering
      \includegraphics[width=\linewidth]{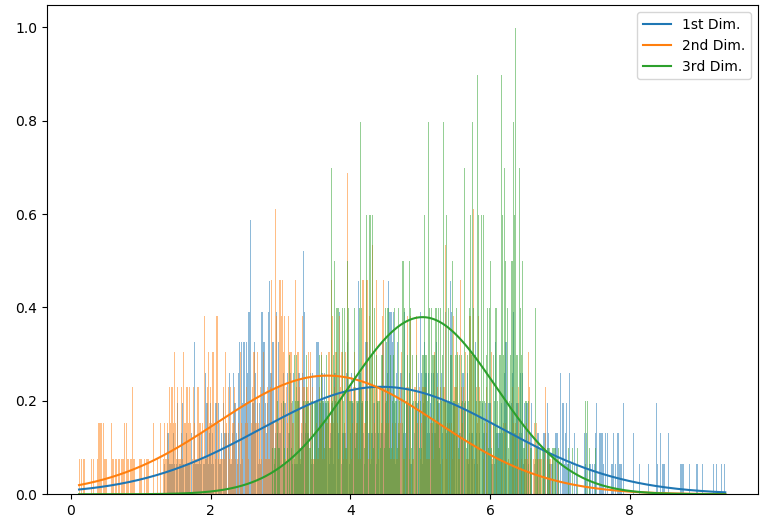}
      \caption{Imagenette Reduction Distributions --- Class `English Springer'}\label{fig:imagenette_distribution_1}
    \end{minipage}
\end{figure}

\begin{figure}[!htbp]
    \centering
    \begin{minipage}[t]{.49\textwidth}
      \centering
      \includegraphics[width=\linewidth]{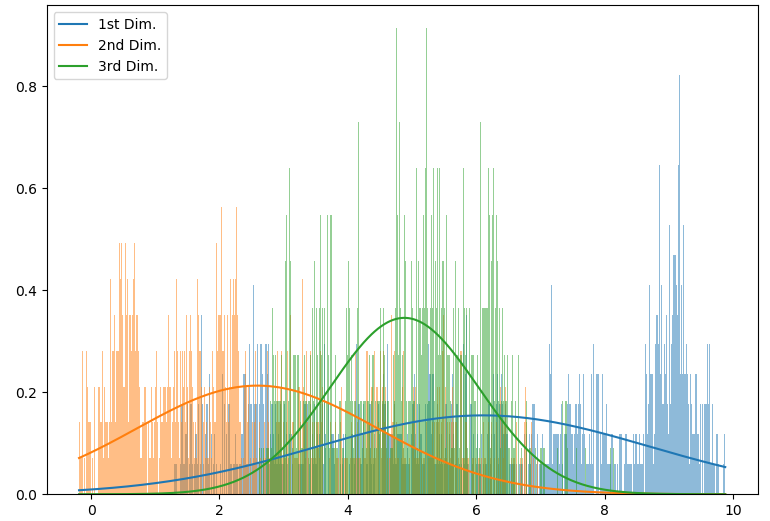}
      \caption{Imagenette Reduction Distributions --- Class `Cassette Player'}\label{fig:imagenette_distribution_2}
    \end{minipage}
    \hfill
    \begin{minipage}[t]{.49\textwidth}
      \centering
      \includegraphics[width=\linewidth]{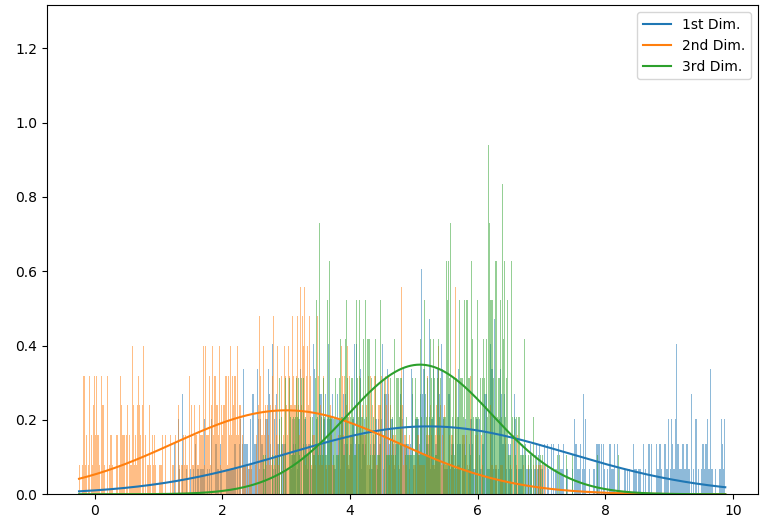}
      \caption{Imagenette Reduction Distributions --- Class `Chain Saw'}\label{fig:imagenette_distribution_3}
    \end{minipage}
\end{figure}

\begin{figure}[!htbp]
    \centering
    \begin{minipage}[t]{.49\textwidth}
      \centering
      \includegraphics[width=\linewidth]{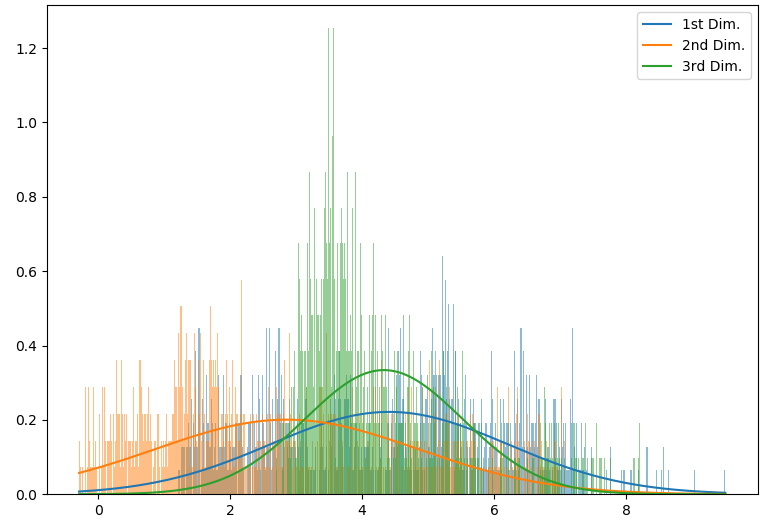}
      \caption{Imagenette Reduction Distributions --- Class `Church'}\label{fig:imagenette_distribution_4}
    \end{minipage}
    \hfill
    \begin{minipage}[t]{.49\textwidth}
      \centering
      \includegraphics[width=\linewidth]{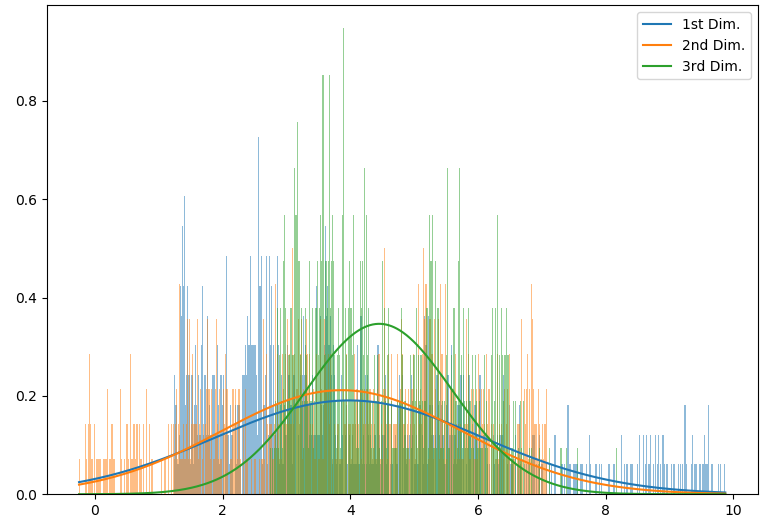}
      \caption{Imagenette Reduction Distributions --- Class `French Horn'}\label{fig:imagenette_distribution_5}
    \end{minipage}
\end{figure}

\begin{figure}[!htbp]
    \centering
    \begin{minipage}[t]{.49\textwidth}
      \centering
      \includegraphics[width=\linewidth]{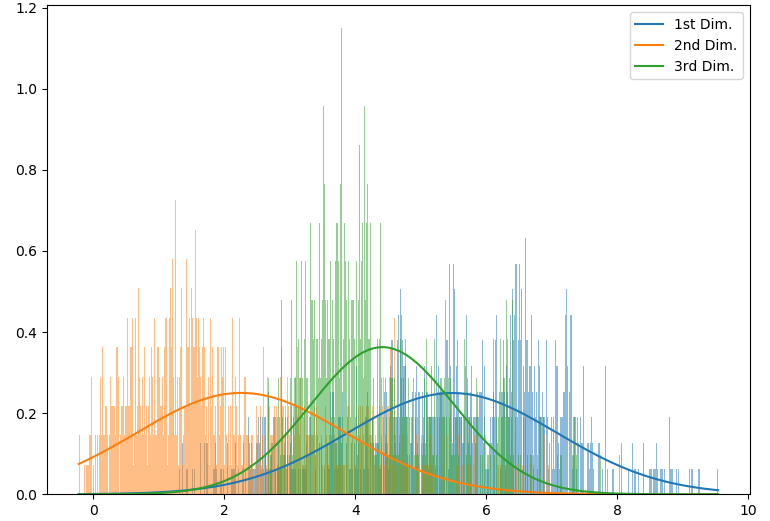}
      \caption{Imagenette Reduction Distributions --- Class `Garbage Truck'}\label{fig:imagenette_distribution_6}
    \end{minipage}
    \hfill
    \begin{minipage}[t]{.49\textwidth}
      \centering
      \includegraphics[width=\linewidth]{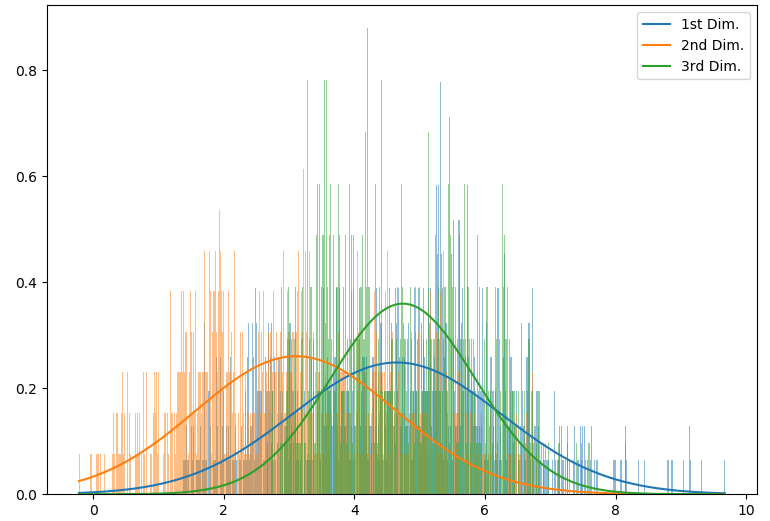}
      \caption{Imagenette Reduction Distributions --- Class `Gas Pump'}\label{fig:imagenette_distribution_7}
    \end{minipage}
\end{figure}

\begin{figure}[!htbp]
    \centering
    \begin{minipage}[t]{.49\textwidth}
      \centering
      \includegraphics[width=\linewidth]{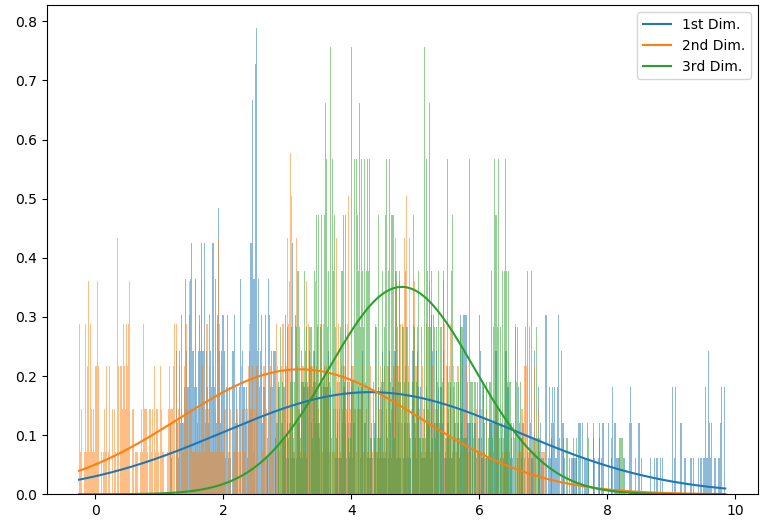}
      \caption{Imagenette Reduction Distributions --- Class `Golf Ball'}\label{fig:imagenette_distribution_8}
    \end{minipage}
    \hfill
    \begin{minipage}[t]{.49\textwidth}
      \centering
      \includegraphics[width=\linewidth]{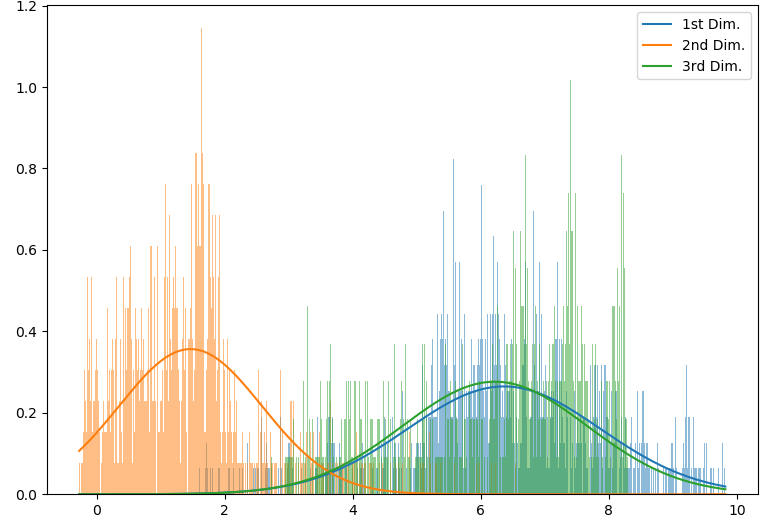}
      \caption{Imagenette Reduction Distributions --- Class `Parachute'}\label{fig:imagenette_distribution_9}
    \end{minipage}
\end{figure}

In \autoref{fig:mnist_distribution_0} through \autoref{fig:imagenette_distribution_9} the values of each of the 3 dimensions of the reductions for each individual class of the training data for MNIST and Imagenette are plotted.  The first thing that can be learned from looking at the MNIST plots is that the reduction for each class produces values significantly different than the others and further, the values in each dimension of each class are tightly grouped in the number line, with the noticeable exception of the third dimension of the class that represents the digit 1.  It is worth noting that the stylization of the Hindu-Arabic numeral '1' contains most of its information in 2-dimensions (accounting for translation) thus providing an explanation for the larger variance in the third dimension.  The plots of the Imagenette classes, on the other hand, are much more similar to one another and display greater variance in each of the 3 dimensions.

\section{Summary}

The experiments we performed show that, for the datasets examined, certain training samples are more informative of class membership than others.  These samples can be identified \textit{a priori} to training by analyzing their position in reduced dimensional space relative to the classes' centroids.  Specifically, we demonstrated that samples nearer the classes' centroids are less informative than those that are furthest from it.  For the five datasets investigated, we have shown that there was no statistically significant difference from the baseline when excluding up to 2\% of the data nearest to each class's centroid.  For CIFAR-100, superior accuracy was achieved when excluding 1\% of the data nearest to each class's centroid.  For CIFAR-10, superior accuracy was achieved when excluding 5\% of the data nearest to each class's centroid.  And for the MNIST dataset, identical accuracy to the baseline was achieved when excluding 10\% of the data nearest to each class's centroid.

Additionally, we defined \textit{Dataset Severability} to be a qualitative, yet quantifiable, judgement regarding the separation of classes in a reduced dimensional space.  High severability was shown for MNIST and Fashion-MNIST, whereas low severability was shown for CIFAR-10, CIFAR-100, and Imagenette.  Those datasets that demonstrated high severability all achieved higher accuracies in the experiments detailed in this work compared to the accuracies of those experiments for the datasets that demonstrated low severability.

\FloatBarrier{}
\printbibliography{}

@article{Maaten2008,
    author = {van der Maaten, Laurens and Hinton, Geoffrey},
    journal = {Journal of Machine Learning Research},
    pages = {2579--2605},
    title = {{Visualizing Data using t-SNE Laurens}},
    volume = {9},
    year = {2008}
}

@article{McInnes2018,
    archivePrefix = {arXiv},
    arxivId = {1802.03426},
    author = {McInnes, Leland and Healy, John and Melville, James},
    title = {{UMAP: Uniform Manifold Approximation and Projection for Dimension Reduction}},
    year = {2018}
}

@article{Byerly2021a,
  author = {Byerly, Adam and Kalganova, Tatiana},
  journal= {IEEE Access},
  title = {Homogeneous Vector Capsules Enable Adaptive Gradient Descent in Convolutional Neural Networks},
  year={2021},
  volume={9},
  pages={48519-48530},
  doi={doi:10.1109/ACCESS.2021.3066842}
}

@article{Byerly2021b,
  title = {No Routing Needed Between Capsules},
  journal = {Neurocomputing},
  volume = {463},
  pages = {545-553},
  year = {2021},
  issn = {0925-2312},
  doi = {https://doi.org/10.1016/j.neucom.2021.08.064},
  author = {Adam Byerly and Tatiana Kalganova and Ian Dear},
}

@inproceedings{Byerly2021c,
  author="Byerly, Adam and Kalganova, Tatiana and Grichnik, Anthony J.",
  title="On the Importance of Capturing a Sufficient Diversity of Perspective for the Classification of Micro-PCBs",
  booktitle="Intelligent Decision Technologies",
  year="2021",
  publisher="Springer Singapore",
  volume = {238},
  pages="209--219",
  isbn="978-981-16-2765-1"
}

@misc{imagenette,
  author    = "Jeremy Howard",
  title     = "Imagenette",
  url       = "https://github.com/fastai/imagenette/",
  year      = {2018},
  urldate   = "2022-01-21"
}

@misc{Xiao2017,
  title = {Fashion-MNIST: a Novel Image Dataset for Benchmarking Machine Learning Algorithms},
  author = {Xiao, Han and Rasul, Kashif and Vollgraf, Roland},
  year = {2017},
  eprint = {1708.07747},
  archivePrefix = {arXiv},
  primaryClass={cs.LG}
}

@article{Lecun2010,
  title={MNIST handwritten digit database},
  author={LeCun, Yann and Cortes, Corinna and Burges, CJ},
  journal={ATT Labs [Online].},
  url={http://yann.lecun.com/exdb/mnist},
  volume={2},
  year={2010}
}

@techreport{Krizhevsky2009,
  title = {Learning Multiple Layers of Features from Tiny Images},
  author = {Krizhevsky, Alex},
  booktitle = {Techincal Report},
  year = {2009}
}

@article{Cover1967,
  author={Cover, T. and Hart, P.},
  journal={IEEE Transactions on Information Theory}, 
  title={Nearest neighbor pattern classification}, 
  year={1967},
  volume={13},
  number={1},
  pages={21-27},
  doi={10.1109/TIT.1967.1053964}
}

@article{Hart1968,
  author={Hart, P.},
  journal={IEEE Transactions on Information Theory},
  title={The condensed nearest neighbor rule (Corresp.)},
  year={1968},
  volume={14},
  number={3},
  pages={515-516},
  doi={doi:10.1109/TIT.1968.1054155}
}

@article{Ritter1975,
  author={Ritter, G. and Woodruff, H. and Lowry, S. and Isenhour, T.},
  journal={IEEE Transactions on Information Theory}, 
  title={An algorithm for a selective nearest neighbor decision rule (Corresp.)}, 
  year={1975},
  volume={21},
  number={6},
  pages={665-669},
  doi={doi:10.1109/TIT.1975.1055464}
}

@article{Wilson1972,
  author={Wilson, Dennis L.},
  journal={IEEE Transactions on Systems, Man, and Cybernetics}, 
  title={Asymptotic Properties of Nearest Neighbor Rules Using Edited Data}, 
  year={1972},
  volume={SMC-2},
  number={3},
  pages={408-421},
  doi={doi:10.1109/TSMC.1972.4309137}
}

@inproceedings{ChienHsing2006,
  author={Chien-Hsing Chou and Bo-Han Kuo and Fu Chang},
  booktitle={18th International Conference on Pattern Recognition (ICPR'06)}, 
  title={The Generalized Condensed Nearest Neighbor Rule as A Data Reduction Method}, 
  year={2006},
  volume={2},
  number={},
  pages={556-559},
  doi={doi:10.1109/ICPR.2006.1119}
}

@inproceedings{Vazquez2005,
  author="V{\'a}zquez, Fernando and S{\'a}nchez, J. Salvador and Pla, Filiberto",
  title="A Stochastic Approach to Wilson's Editing Algorithm",
  booktitle="Pattern Recognition and Image Analysis",
  year="2005",
  pages="35--42",
  isbn="978-3-540-32238-2"
}

@article{Wilson2000,
  author = {Wilson, D Randall and Martinez, Tony R},
  journal = {Machine Learning},
  pages = {257--286},
  title = {{Reduction Techniques for Instance-Based Learning Algorithms}},
  volume = {38},
  year = {2000}
}

@phdthesis{Albalate2007,
  author = {Albalate, Mar{\'i}a Teresa Lozano},
  title = {{Data reduction techniques in classification processes}},
  year = {2007}
}

@inproceedings{Ougiaroglou2012,
  author = {Ougiaroglou, Stefanos and Evangelidis, Georgios},
  booktitle = {Balkan Conference in Informatics (BCI)},
  doi = {doi:10.1145/2371316.2371349},
  isbn = {9781450312400},
  pages = {168--173},
  title = {{Efficient dataset size reduction by finding homogeneous clusters}},
  year = {2012}
}

@inproceedings{Krizhevsky2012,
  author = {Krizhevsky, Alex and Sutskever, Ilya and Hinton, Geoffrey E},
  booktitle = {NIPS 2012 - 25th Conference on Neural Information Processing Systems},
  doi = {doi:10.1145/3065386},
  pages = {1097--1105},
  title = {{ImageNet Classification with Deep Convolutional Neural Networks}},
  year = {2012}
}

@article{Shayegan2014,
  author = {Shayegan, Mohammad Amin and Aghabozorgi, Saeed},
  doi = {doi:10.1155/2014/537428},
  issn = {15635147},
  journal = {Mathematical Problems in Engineering},
  title = {{A new dataset size reduction approach for PCA-based classification in OCR application}},
  year = {2014}
}

@article{Zhai2021,
    archivePrefix = {arXiv},
    arxivId = {2106.04560},
    author = {Zhai, Xiaohua and Kolesnikov, Alexander and Houlsby, Neil and Beyer, Lucas},
    title = {{Scaling Vision Transformers}},
    url = {http://arxiv.org/abs/2106.04560},
    year = {2021}
}

@article{Riquelme2021,
    archivePrefix = {arXiv},
    arxivId = {2106.05974},
    author = {Riquelme, Carlos and Puigcerver, Joan and Mustafa, Basil and Neumann, Maxim and Jenatton, Rodolphe and Pinto, Andr{\'{e}} Susano and Keysers, Daniel and Houlsby, Neil},
    title = {{Scaling Vision with Sparse Mixture of Experts}},
    url = {http://arxiv.org/abs/2106.05974},
    year = {2021}
}

@article{Ryoo2021,
    archivePrefix = {arXiv},
    arxivId = {2106.11297},
    author = {Ryoo, Michael S. and Piergiovanni, AJ and Arnab, Anurag and Dehghani, Mostafa and Angelova, Anelia},
    title = {{TokenLearner: What Can 8 Learned Tokens Do for Images and Videos?}},
    url = {http://arxiv.org/abs/2106.11297},
    year = {2021}
}

@inproceedings{Jia2021,
    author = {Jia, Chao and Yang, Yinfei and Xia, Ye and Chen, Yi-Ting and Parekh, Zarana and Pham, Hieu and Le, Quoc V. and Sung, Yunhsuan and Li, Zhen and Duerig, Tom},
    booktitle = {Proceedings of the 38th International Conference on Machine Learning (PMLR)},
    title = {{Scaling Up Visual and Vision-Language Representation Learning With Noisy Text Supervision}},
    year = {2021}
}

@inproceedings{Dosovitskiy2020,
    author = {Dosovitskiy, Alexey and Beyer, Lucas and Kolesnikov, Alexander and Weissenborn, Dirk and Zhai, Xiaohua and Unterthiner, Thomas and Dehghani, Mostafa and Minderer, Matthias and Heigold, Georg and Gelly, Sylvain and Uszkoreit, Jakob and Houlsby, Neil},
    booktitle = {Ninth International Conference on Learning Representations (ICLR)},
    title = {{An Image is Worth 16x16 Words: Transformers for Image Recognition at Scale}},
    year = {2020}
}

@inproceedings{Kolesnikov2020,
    author = {Kolesnikov, Alexander and Beyer, Lucas and Zhai, Xiaohua and Puigcerver, Joan and Yung, Jessica and Gelly, Sylvain and Houlsby, Neil},
    booktitle = {16th European Conference on Computer Vision},
    title = {{Big Transfer (BiT): General Visual Representation Learning}},
    year = {2020}
}

@article{Dong2021,
    archivePrefix = {arXiv},
    arxivId = {2107.00652},
    author = {Dong, Xiaoyi and Bao, Jianmin and Chen, Dongdong and Zhang, Weiming and Yu, Nenghai and Yuan, Lu and Chen, Dong and Guo, Baining},
    title = {{CSWin Transformer: A General Vision Transformer Backbone with Cross-Shaped Windows}},
    url = {http://arxiv.org/abs/2107.00652},
    year = {2021}
}

@inproceedings{Liu2021,
    author = {Liu, Ze and Lin, Yutong and Cao, Yue and Hu, Han and Wei, Yixuan and Zhang, Zheng and Lin, Stephen and Guo, Baining},
    booktitle = {The International Conference on Computer Vision (ICCV)},
    title = {{Swin Transformer: Hierarchical Vision Transformer using Shifted Windows}},
    year = {2021}
}

@article{Dai2021,
    archivePrefix = {arXiv},
    arxivId = {2106.04803},
    author = {Dai, Zihang and Liu, Hanxiao and Le, Quoc V. and Tan, Mingxing},
    title = {{CoAtNet: Marrying Convolution and Attention for All Data Sizes}},
    url = {http://arxiv.org/abs/2106.04803},
    year = {2021}
}

@inproceedings{Wu2021,
    author = {Wu, Haiping and Xiao, Bin and Codella, Noel and Liu, Mengchen and Dai, Xiyang and Yuan, Lu and Zhang, Lei},
    booktitle = {The International Conference on Computer Vision (ICCV)},
    title = {{CvT: Introducing Convolutions to Vision Transformers}},
    year = {2021}
}

@article{Touvron2020a,
    author = {Touvron, Hugo and Vedaldi, Andrea and Douze, Matthijs and J{\'{e}}gou, Herv{\'{e}}},
    journal = {Advances in Neural Information Processing Systems},
    title = {{Fixing the train-test resolution discrepancy: FixEfficientNet}},
    volume = {32},
    year = {2019}
}

@article{Tan2021,
    archivePrefix = {arXiv},
    arxivId = {2104.00298},
    author = {Tan, Mingxing and Le, Quoc V.},
    title = {{EfficientNetV2: Smaller Models and Faster Training}},
    url = {http://arxiv.org/abs/2104.00298},
    year = {2021}
}

@article{Tolstikhin2021,
    archivePrefix = {arXiv},
    arxivId = {2105.01601},
    author = {Tolstikhin, Ilya and Houlsby, Neil and Kolesnikov, Alexander and Beyer, Lucas and Zhai, Xiaohua and Unterthiner, Thomas and Yung, Jessica and Steiner, Andreas and Keysers, Daniel and Uszkoreit, Jakob and Lucic, Mario and Dosovitskiy, Alexey},
    title = {{MLP-Mixer: An all-MLP Architecture for Vision}},
    url = {http://arxiv.org/abs/2105.01601},
    year = {2021}
}

@article{Pham2020,
    archivePrefix = {arXiv},
    arxivId = {2003.10580},
    author = {Pham, Hieu and Dai, Zihang and Xie, Qizhe and Luong, Minh-Thang and Le, Quoc V.},
    title = {{Meta Pseudo Labels}},
    url = {http://arxiv.org/abs/2003.10580},
    year = {2020}
}

@inproceedings{Xie2020,
    author = {Xie, Qizhe and Luong, Minh Thang and Hovy, Eduard and Le, Quoc V.},
    doi = {doi:10.1109/CVPR42600.2020.01070},
    issn = {10636919},
    booktitle = {Proceedings of the IEEE Computer Society Conference on Computer Vision and Pattern Recognition},
    pages = {10684--10695},
    title = {{Self-training with noisy student improves imagenet classification}},
    year = {2020}
}

@article{Brock2021,
    archivePrefix = {arXiv},
    arxivId = {2102.06171},
    author = {Brock, Andrew and De, Soham and Smith, Samuel L. and Simonyan, Karen},
    title = {{High-Performance Large-Scale Image Recognition Without Normalization}},
    url = {http://arxiv.org/abs/2102.06171},
    year = {2021}
}

@inproceedings{Foret2020,
    author = {Foret, Pierre and Kleiner, Ariel and Mobahi, Hossein and Neyshabur, Behnam},
    booktitle = {Ninth International Conference on Learning Representations (ICLR)},
    title = {{Sharpness-Aware Minimization for Efficiently Improving Generalization}},
    year = {2020}
}

\end{document}